\documentclass{article}

\usepackage[final]{neurips_2022}
\usepackage[utf8]{inputenc}
\usepackage[T1]{fontenc}
\usepackage{hyperref}
\usepackage{url}
\usepackage{booktabs}
\usepackage{amsfonts}
\usepackage{nicefrac}
\usepackage{microtype}
\usepackage{xcolor}
\usepackage{algorithm}
\usepackage[noend]{algorithmic}
\usepackage{subcaption}
\usepackage{microtype}
\usepackage{graphicx}
\usepackage{booktabs} 
\usepackage{hyperref}
\usepackage{amsmath}
\usepackage{amssymb}
\usepackage{bm}
\usepackage{mathtools}
\usepackage{amsthm}
\usepackage{xspace}
\usepackage[textsize=tiny]{todonotes}
\usepackage{bbm}
\usepackage[capitalize,noabbrev]{cleveref}

\theoremstyle{plain}
\newtheorem{theorem}{Theorem}[section]
\newtheorem{proposition}[theorem]{Proposition}

\newtheorem{corollary}[theorem]{Corollary}
\theoremstyle{definition}

\theoremstyle{remark}

\newcommand{\kl}{k_{\lambda}}
\newcommand{\bkl}{{\bar k}_{\lambda}}
\newcommand{\hl}{h_{p,\kl}}
\newcommand{\hh}{h_{p,k}}
\newcommand{\vl}{\varphi_\lambda}
\newcommand{\db}{\Delta_{\alpha}^{k}}
\newcommand{\dbv}{\Delta_{\alpha}^{k}(\Xn)}
\newcommand{\dbvll}{\Delta_{\alpha}^{\lambda}(\Xn)}
\newcommand{\dbu}{\Delta_{u_\alpha w_k}^{k}}
\newcommand{\dbuv}{\Delta_{u_\alpha w_k}^{k}(\Xn)}
\newcommand{\dbb}{\Delta_{\alpha}^{\K}}
\newcommand{\dbbv}{\Delta_{\alpha}^{\K}(\Xn)}
\newcommand{\dbbvll}{\Delta_{\alpha}^{\Lambda}(\Xn)}
\newcommand{\dbbl}{\Delta_{\alpha}^{\K_\Lambda}}
\newcommand{\dbblv}{\Delta_{\alpha}^{\K_\Lambda}(\Xn)}

\newcommand{\ksdv}{\mathrm{KSD}_{p,k}^2(q)}

\newcommand{\ksdh}{\widehat{\mathrm{KSD}}_{p,k}^2}
\newcommand{\ksdhv}{\widehat{\mathrm{KSD}}_{p,k}^2(\Xn)}

\newcommand{\ksdagg}{{\sc KSDAgg}\xspace}
\newcommand{\Sb}{\mathcal{S}_d^s(R)}
\newcommand{\Sbt}{\mathcal{S}_d^{s,t}(R, L)}
\newcommand{\p}[1]{\!\left( #1 \right)}
\newcommand{\R}{\mathbb R}
\newcommand{\E}{\mathbb E}
\newcommand{\K}{\mathcal K}

\newcommand{\HH}{\mathcal H}
\newcommand{\Xn}{\mathbb{X}_N}

\newcommand{\dd}{\mathrm{d}}
\newcommand{\qq}{\widehat{q}_{1-\alpha}^{\,k}}

\newcommand{\qqi}{\widehat{q}_{1-\alpha}^{\,k,\infty}}

\newcommand{\one}[1]{\mathbbm{1}\!\left(#1\right)}

\newcommand{\N}{{\mathbb N}}
\newcommand{\pr}[1]{{{\mathbb P}_{p}\!\left(#1\right)}}
\newcommand{\pe}[1]{{{\mathbb P}_{\epsilon}\!\left(#1\right)}}
\newcommand{\qr}[1]{{{\mathbb P}_{q}\!\left(#1\right)}}

\newcommand{\aamin}[1]{\underset{#1}{\mathrm{min}\, }}
\newcommand{\aamax}[1]{\underset{#1}{\mathrm{max}\, }}

\DeclarePairedDelimiter\abs{\lvert}{\rvert}
\DeclarePairedDelimiter\norm{\lVert}{\rVert}
\makeatletter
\let\oldabs\abs
\def\abs{\@ifstar{\oldabs}{\oldabs*}}
\let\oldnorm\norm
\def\norm{\@ifstar{\oldnorm}{\oldnorm*}}
\makeatother

\DeclarePairedDelimiter{\ceil}{\lceil}{\rceil}

\title{KSD Aggregated Goodness-of-fit Test}
\author{%
Antonin Schrab \\
Centre for Artificial Intelligence\\
Gatsby Computational Neuroscience Unit\\
University College London \& Inria London \\ 
\texttt{a.schrab@ucl.ac.uk}
\And
Benjamin Guedj \\
Centre for Artificial Intelligence\\
University College London \& Inria London \\ 
\texttt{b.guedj@ucl.ac.uk} \\
\And
Arthur Gretton \\
Gatsby Computational Neuroscience Unit\\
University College London \\
\texttt{arthur.gretton@gmail.com}
}

\begin{document}

\maketitle

\begin{abstract}
    We investigate properties of goodness-of-fit tests based on the Kernel Stein Discrepancy (KSD).
    We introduce a strategy to construct a test, called \mbox{\ksdagg}, which aggregates multiple tests with different kernels.
    \mbox{\ksdagg} avoids splitting the data to perform kernel selection (which leads to a loss in test power), and rather maximises the test power over a collection of kernels.
    We provide non-asymptotic guarantees on the power of \mbox{\ksdagg}: we show it achieves the smallest uniform separation rate of the collection, up to a logarithmic term.
    For compactly supported densities with bounded model score function, we derive the rate for \mbox{\ksdagg} over restricted Sobolev balls; this rate corresponds to the minimax optimal rate over unrestricted Sobolev balls, up to an iterated logarithmic term.
    \mbox{\ksdagg} can be computed exactly in practice as it relies 
    either on a parametric bootstrap or on a wild bootstrap
    to estimate the quantiles and the level corrections.
    In particular, for the crucial choice of bandwidth of a fixed kernel, it avoids resorting to arbitrary heuristics (such as median or standard deviation) or to data splitting.
    We find on both synthetic and real-world data that \mbox{\ksdagg} outperforms other state-of-the-art quadratic-time adaptive KSD-based goodness-of-fit testing procedures.
\end{abstract}

\section{Introduction}
    \label{sec:intoduction}

    Kernel selection remains a fundamental problem in kernel-based nonparametric hypothesis testing, as it significantly impacts the test power. 
    Kernel selection has attracted a significant interest in the literature, and a number of methods have been proposed in the two-sample, independence and goodness-of-fit testing frameworks, such as using heuristics \citep{gretton2012kernel}, relying on data splitting \citep{gretton2012optimal,sutherland2016generative,kubler2022witness}, learning deep kernels \citep{grathwohl2020learning,liu2020learning}, working in the post-selection inference framework \citep{yamada2018post,lim2019kernel,lim2020more,kubler2020learning,freidling2021post}, to name but a few.

    In this work, we focus on aggregated tests, which have been
    investigated for the two-sample problem by \citet{fromont2013two}, \citet{kim2020minimax} and \citet{schrab2021mmd} using the Maximum Mean Discrepancy \citep[MMD;][]{gretton2012kernel}, and for the independence problem by \citet{albert2019adaptive} and \citet{kim2020minimax} using the Hilbert Schmidt Independence Criterion \citep[HSIC;][]{gretton2005measuring}.
    We extend the use of aggregated tests to the goodness-of-fit setting,
    where we are given a model and some samples, and test whether the samples have been drawn from the model.
    We employ the Kernel Stein Discrepancy \citep[KSD;][]{chwialkowski2016kernel,liu2016kernelized} as our test statistic,
    which is an ideal measure of distance for this setting: it admits an estimator which can be computed without requiring samples from the model,
    and does not require the model to be normalised.
    To the best of our knowledge, ours represents the first aggregation procedure for the KSD test in the literature.

    \paragraph{Related work.} 
    \citet{fromont2012kernels,fromont2013two} introduced non-asymptotic aggregated tests for the two-sample problem with sample sizes following a Poisson process, using an unscaled version of the MMD. 
    Using a wild bootstrap, they derived uniform separation rates \citep{ingster1987minimax,ingster1999asymptotically,ingster1993asymptotically,ingster1993minimax,baraud2002non}.
    \citet{albert2019adaptive} then proposed an independence aggregated test using the HSIC, with guarantees using a theoretical quantile, but relying on permutations to obtain the test threshold in practice.
    \citet{kim2020minimax} then extended those theoretical results to also hold for the quantile estimated using permutations, however, they did not obtain the desired level dependency in their HSIC bound.
    All those aforementioned results were proved for the Gaussian kernel only.
    \citet{schrab2021mmd} generalised the two-sample results to hold for the usual MMD estimator and for a wide range of kernels using either a wild bootstrap or permutations, and provided optimality results which hold with fewer restrictions.
    Our work builds and extends on the above non-asymptotic results: we consider the goodness-of-fit framework, where we have samples from only one of the two densities. 
    The main challenges arise from working with the Stein kernel which defines the KSD test statistic: for example, we lose the transition-invariant property of the kernel which is crucial to work in the Fourier domain.
    \citet{balasubramanian2021optimality} considered adaptive MMD-based goodness-of-fit tests and obtained their uniform separation rates over Sobolev balls in the asymptotic regime.
    More generally, \citet{li2019optimality} studied asymptotic adaptive kernel-based tests for the three testing frameworks.
    \citet{tolstikhin2016minimax} derived minimax rates for MMD estimators using radial universal kernel \citep{sriperumbudur2011universality}.
    \cite{schrab2022efficient} extends this work, together with those of \citet{albert2019adaptive} and \citet{schrab2021mmd}, to construct efficient aggregated tests for the three testing frameworks using incomplete $U$-statistics. 
    They quantify the cost in the minimax rate over Sobolev balls incurred for computational efficiency, and prove minimax optimality of the quadratic-time HSIC permuted aggregated test by improving the bound of \citet{kim2020minimax}.
    See \Cref{sec:appendix_background} for details.

    \paragraph{Contributions.} 
    We propose a solution to the fundamental kernel selection problem for the widely-used KSD goodness-of-fit tests: we construct an adaptive test \ksdagg which aggregates multiple tests with different kernels. 
    Our contribution is in showing, both theoretically and experimentally, that the aggregation procedure works in this novel setting in which it has never been considered before.
    We work in the kernel selection framework; this general setting has many applications including the one of kernel bandwidth selection.
    Our aggregated test allows for two numerical methods for estimating the test thresholds: the wild bootstrap and the parametric bootstrap (a procedure unique to the goodness-of-fit framework).
    We conduct a theoretical analysis: we derive a general condition which guarantees test power for \ksdagg in terms of its uniform separation rate, with extra assumptions including regularity over restricted Sobolev balls we prove that \ksdagg attains the minimax rate over (unrestricted) Sobolev balls.
    We discuss the implementation of \ksdagg and experimentally validate our proposed approach on benchmark problems, not only on datasets classically used in the literature but also on original data obtained using state-of-the-art generative models (\emph{i.e.} Normalizing Flows).
    We observe, both on synthetic and real-world data, that \ksdagg obtains higher power than other KSD-based adaptive state-of-the-art tests.
    Contributing to the real-world applications of these goodness-of-fit tests, we provide publicly available code to allow practitioners to employ our method: \url{https://github.com/antoninschrab/ksdagg-paper}.

\section{Notation}
    \label{sec:background}

    We consider the goodness-of-fit problem where given access to a known model probability density $p$ (which can be unnormalised since we actually only need access to its score function $\nabla \log p(\cdot)$) and to some i.i.d. $d$-dimensional samples $\Xn \coloneqq (X_i)_{i=1}^N$ drawn from an unknown data density $q$, we want to decide whether $p\neq q$ holds.
    This can be expressed as a statistical hypothesis testing problem with null hypothesis $\HH_0:p=q$ and alternative $\HH_a:p\neq q$.

    As a measure of distance between $p$ and $q$, we use the {\em Kernel Stein Discrepancy} (KSD) introduced by \citet{chwialkowski2016kernel} and \citet{liu2016kernelized}. 
    For a kernel $k$, the KSD is the Maximum Mean Discrepancy \citep[MMD;][]{gretton2012kernel} between $p$ and $q$ using the Stein kernel associated to $k$
    \begin{align*}
        \mathrm{KSD}_{p,k}^2(q) 
        \coloneqq \mathrm{MMD}^2_{\hh}(p,q)
        \coloneqq\ &\E_{q,q}[\hh(X,Y)] 
        - 2 \E_{p,q}[\hh(X,Y)]+ \E_{p,p}[\hh(X,Y)] \\
        =\ &\E_{q,q}[\hh(X,Y)]
    \end{align*}
    where the {\em Stein kernel} $\hh\colon\R^d\times\R^d\to\R$ is defined as
    \begin{align*}
        \hh(x,y) \coloneqq\ 
        &\p{\nabla\log p(x)^\top \nabla\log p(y)} k(x,y)
        + \nabla\log p(y)^\top \nabla_x k(x,y) \\
        &+ \nabla\log p(x)^\top \nabla_y k(x,y) 
        + \sum_{i=1}^d \frac{\partial}{\partial x_i \partial y_i}\, k(x,y)
    \end{align*}
    and satisfies the {\em Stein identity}
    $
        \E_{p}[\hh(X,\cdot)] = 0
    $.
    Additional background details on the KSD are presented in \Cref{sec:appendix_background}.
    A quadratic-time {\em KSD estimator} can be computed as the $U$-statistic \citep{hoeffding1992class}
    \begin{equation}
        \label{eq:ksdh}
        \ksdhv 
        \coloneqq \frac{1}{N(N-1)}\sum_{1\leq i\neq j \leq N}\hh(X_i,X_j).
    \end{equation}
    In this work, the score of the model density $p$ is always known, we do not always explicitly write the dependence on $p$ of all the variables we consider.
    We assume that the kernel $k$ is such that
    \begin{equation}
        \label{eq:ksd}
        \ksdv =\E_{q,q}[\hh(X,Y)] < \infty
        \qquad  \textrm{and} \qquad
        C_k \coloneqq \E_{q,q}[\hh(X,Y)^2] < \infty.
    \end{equation}    
    We now address the requirements for consistency (\emph{i.e.} test power converges to 1 as the sample size goes to $\infty$) of the Stein test \citep[Theorem 2.2]{chwialkowski2016kernel}: we assume that the kernel $k$ is $C_0$-universal \citep[Definition 4.1]{carmeli2010vector} and that $\mathbb{E}_q \Big[\norm{\nabla \p{\log \frac{p(X)}{q(X)}}}_2^2\Big]<\infty$.

    We use the notations $\mathbb{P}_p$ and $\mathbb{P}_q$ to denote the probability under the model distribution $p$ and under the data distribution $q$, respectively.
    Given a kernel $\kappa\colon\R^d\times\R^d\to\R$ and a function $f\colon\R^d\to\R$ in $L^2\p{\R^d}$,
    we consider the {\em integral transform} $T_\kappa$ defined as
    \begin{align*}
        \p{T_\kappa f}(y) 
        \coloneqq \int_{\R^d} \kappa(x,y) f(x) \,\dd x
    \end{align*}
    for $y\in\R^d$.
    When the kernel $\kappa$ is translation-invariant, the integral transform corresponds to a convolution. 
    However, the lack of translation invariance of the Stein kernel introduces new challenging problems. 
    First, working the integral transform of the Stein kernel is more complicated since it does not correspond to a simple convolution.
    Second, for the expectation of the Stein kernel squared, $C_k \coloneqq \E_{q,q}[\hh(X,Y)^2]$, it is not possible to extract the bandwidth parameter $\lambda$ outside of the expectation as it is the case when using the usual kernel directly as for the MMD and HSIC.

\section{Construction of tests and bounds}
    \label{sec:theory}

    We now introduce the single and aggregated KSD tests. 
    We show that these control the probability of type I error as desired, and provide conditions for the control of the probability of type II error.

\subsection{Single test}
    \label{subsec:singletest}

    We first construct a KSD test for a fixed kernel $k$ as proposed by \citet{chwialkowski2016kernel} and \citet{liu2016kernelized}. 
    To estimate the test threshold, we can either use a wild bootstrap \citep{shao2010dependent,leucht2013dependent,fromont2012kernels,chwialkowski2014wild} or a parametric bootstrap \citep{stute1993bootstrap}.
    Both methods work by simulating sampling values $\big(\bar{K}_{k}^{1}, \dots, \bar{K}_{k}^{B_1}\big)$ from the (asymptotic) distribution of $\ksdh$ under the null hypothesis and estimating the $(1\!-\!\alpha)$-quantile for $\alpha\in(0,1)$ using a Monte Carlo approximation\footnote{We do not write explicitly the dependence of $\qq$ on other variables, but those are implicitly considered when writing probabilistic statements.}
    \begin{equation}
        \label{eq:quantileq}
        \qq
        \coloneqq\ \inf\!\bigg\{u \in \R: 1 - \alpha \leq \frac{1}{B_1+1}\sum_{b=1}^{B_1+1} \one{\bar K_{k}^b\leq u}\!\bigg\}
        = \bar{K}_{k}^{\,\bullet\ceil{(B_1+1)(1-\alpha)}} 
    \end{equation}
    where $\bar{K}_{k}^{\bullet 1} \leq \dots \leq \bar{K}_{k}^{\bullet B_1+1}$ are the sorted elements $\big(\bar{K}_{k}^{1}, \dots, \bar{K}_{k}^{B_1+1}\big)$ with $\bar{K}_{k}^{B_1+1}\coloneqq \ksdhv$.
    The single test is then defined as
    (a test function outputs 1 when the null is rejected and 0 otherwise)
    $$
        \dbv \coloneqq \one{\ksdhv > \qq}.
    $$

    For the {\em parametric bootstrap}, we directly draw new samples $(X_i')_{i=1}^{N}$ from the model distribution $p$ (it might not always be possible to so) and compute the KSD 
    \begin{equation}
        \label{eq:parametricbootstrap}
        \bar{K}_{k}
        \coloneqq \frac{1}{N(N-1)}\sum_{1\leq i\neq j \leq N} \hh(X_i',X_j').
    \end{equation}
    For the {\em wild bootstrap}, we first generate $n$ i.i.d. Rademacher random variables $\epsilon_1,\dots,\epsilon_n$, each taking value in $\{-1,1\}$, and then compute
    \begin{equation}
        \label{eq:wildbootstrap}
        \bar{K}_{k}
        \coloneqq \frac{1}{N(N-1)}\sum_{1\leq i\neq j \leq N} \epsilon_i \epsilon_j \hh(X_i,X_j).
    \end{equation}
    By repeating either procedure $B_1$ times, we obtain the bootstrapped samples $\bar{K}_{k}^{1}, \dots, \bar{K}_{k}^{B_1}$.

    Since it uses samples from the model $p$, the parametric bootstrap \citep{stute1993bootstrap} results in a test with non-asymptotic level $\alpha$. This comes at the cost of being computationally more expensive and assuming that we are able to sample from $p$ (which may be out of reach in some settings).
    Conversely, the wild bootstrap has the advantage of not requiring to sample from $p$, which makes it computationally more efficient as only one kernel matrix needs to be computed, but it only achieves the desired level $\alpha$ asymptotically \citep{shao2010dependent,leucht2013dependent,leucht2012degenerate,chwialkowski2014wild,chwialkowski2016kernel} assuming Lipschitz continuity of $h_{p,k}$ (see \Cref{sec:wilbootstrap} for details).
    Note that we cannot obtain a non-asymptotic level for the wild bootstrap by relying on the result of \citet[Lemma 1]{romano2005exact} as done in the two-sample framework by \citet{fromont2013two} and \citet{schrab2021mmd}.
    This is because in our case $\bar{K}_{k}$ and $\ksdhv$ are not exchangeable variables under the null hypothesis, due to
    the asymmetry of the KSD statistic with respect to $p$ and $q$.
    
    Having discussed control of the probability of type I error of the single test $\db$, we now provide a condition on $\norm{p-q}_2$ which ensures that the probability of type II error is controlled by some $\beta\in(0,1)$.
    The smallest such value of $\norm{p-q}_2$, provided that $p-q$ lies in some given class of regular functions, is called the {\em uniform separation rate} \citep{ingster1987minimax,ingster1999asymptotically,ingster1993asymptotically,ingster1993minimax,baraud2002non}.

    \begin{theorem}
        \label{lem:singlepower}
        Suppose the assumptions listed in \Cref{assump:singlepower} hold, and let $\psi\coloneqq p-q$.
        There exists a positive constant $C$ such that the condition
        $$
            \norm{\psi}_2^2 
            ~\geq~ \norm{\psi-T_{\hh}\psi}_2^2
            + C \log\p{\frac{1}{\alpha}}\!\frac{\sqrt{C_k}}{\beta N}
        $$
        guarantees control over the probability of type II error, such that
        $
            \qr{\dbv=0} \leq \beta
        $.
    \end{theorem}
    \Cref{lem:singlepower}, which is proved in \Cref{proof:singlepower}, provides a power guaranteeing condition consisting of two terms. 
    The first term $\|\psi-T_{h_{p,k}} \psi\|_2^2$ indicates the size of the effect of the Stein integral transform operator on the difference in densities $\psi \coloneqq p-q$, it is a measure of distance from the null (where this quantity is zero).
    The second term $C\log\!\big(1/\alpha\big) (\beta N)^{-1}\sqrt{C_{k}}$ is obtained from upper bounding the variance of the KSD $U$-statistic, it depends on the expectation of the squared Stein kernel $C_k \coloneqq \mathbb{E}_{q,q}[h_{p,k}(X,Y)^2]$. 
    This second term also controls the quantile of the test.

\subsection{Aggregated test}
    \label{subsec:aggtest}

    We can now introduce our aggregated test, which is motivated by the earlier works of \citet{fromont2012kernels,fromont2013two}, \citet{albert2019adaptive}, and \citet{schrab2021mmd} for two-sample and independence testing.
    
    We compute $\widetilde K_k^1,\dots,\widetilde K_k^{B_2}$ further KSD values simulated from the null hypothesis obtained using either a parametric bootstrap or a wild bootstrap as in Equations (\ref{eq:parametricbootstrap}) or (\ref{eq:wildbootstrap}), respectively.
    We consider a finite collection $\K$ of kernels  satisfying the properties presented in \Cref{sec:background}.
    We construct an aggregated test $\dbb$, called \ksdagg, which rejects the null hypothesis if one of the single tests $\p{\dbu}_{k\in\K}$ rejects it, that is
    \begin{equation*}
        \dbbv 
        \coloneqq \one{\dbuv=1 \textrm{ for some } k\in\K}.    
    \end{equation*}
    The levels of the single tests are adjusted to ensure the aggregated test has the prescribed level $\alpha$. 
    This adjustment is performed by introducing positive weights $(w_k)_{k\in\K}$ satisfying $\sum_{k\in\K} w_k \leq 1$ and some correction
    \begin{equation}
        \label{eq:ualpha}
        u_\alpha
        \coloneqq \sup\!\bigg\{u\in\Big(0,\ \aamin{k\in\K}w_k^{-1}\Big): \widehat P_u \leq \alpha\bigg\}
    \end{equation}
    where 
    $$
        \widehat P_u \coloneqq \frac{1}{B_2}\sum_{b=1}^{B_2}\one{\aamax{k\in\K}\p{\widetilde K_k^b-\bar{K}_{k}^{\,\bullet\ceil{(B_1+1)(1-u w_k)}}}>0}
    $$
    is a Monte Carlo approximation of the type I error probability of the aggregated test with correction $u$
    $$
        P_u \coloneqq \pr{\aamax{k\in\K}\p{\ksdhv-\widehat q_{1-u w_k }^{\,k}}>0}.
    $$
    To compute $u_\alpha$, we estimate the supremum in \Cref{eq:ualpha} by performing $B_3$ steps of the bisection method, the theoretical results account for this extra approximation.
    Detailed pseudocode for \ksdagg is provided in \Cref{alg:ksdagg}, and details regarding our aggregation procedure are provided in \Cref{sec:aggdetails}. We discuss potential limitations of KSDAgg in \Cref{sec:limitations}.

    \begin{algorithm}[t]
       \caption{\ksdagg}
       \label{alg:ksdagg}
    \begin{algorithmic}
        \STATE {\bfseries Inputs:} 
        samples $\Xn=(x_i)_{i=1}^N$, 
        density $p$ or score $\nabla \log p(\cdot)$,
        finite kernel collection $\K$,
        weights $(w_k)_{k\in\K}$,
        level $\alpha\in(0,e^{-1})$,
        estimation parameters $B_1,B_2,B_3\in\N$,
        parametric or wild bootstrap
        
        \STATE {\bfseries Output:} $0$ (fail to reject $\mathcal{H}_0$) or $1$ (reject $\mathcal{H}_0$)
        \STATE {\bfseries Algorithm:} 
        \FOR{$k\in\K$}
        \STATE compute $\bar{K}_{k}^{B_1+1}\coloneqq \ksdhv$ as in \Cref{eq:ksdh}
        \STATE compute $\big(\bar{K}_{k}^{b}\big)_{1\leq b\leq B_1}$ as in Equations \eqref{eq:parametricbootstrap} or \eqref{eq:wildbootstrap}
        \STATE sort $\big(\bar{K}_{k}^{b}\big)_{1\leq b\leq B_1+1}$ in ascending order to obtain $\big(\bar{K}_{k}^{\bullet b}\big)_{1\leq b\leq B_1+1}$
        \STATE compute $\big(\widetilde{K}_{k}^{b}\big)_{1\leq b\leq B_2}$ as in Equations \eqref{eq:parametricbootstrap} or \eqref{eq:wildbootstrap}
        \ENDFOR
        \STATE $u_{\textrm{min}} = 0$, $\ u_{\textrm{max}} = \aamin{k\in\K} w_k^{-1}$
        \FOR{$t=1,\dots, B_3$}
        \STATE $u =  \frac{1}{2}\p{u_{\textrm{min}}+u_{\textrm{max}}}$, 
        $\ \widehat P_u = \displaystyle\frac{1}{B_2}\sum_{b=1}^{B_2}\one{\aamax{k\in\K}\p{\!\widetilde K_k^b-\bar{K}_{k}^{\,\bullet\ceil{(B_1+1)(1-u w_k)}}}>0}$
        \STATE \textbf{if } $\widehat{P}_u\leq \alpha$ \textbf{ then } $u_{\textrm{min}} = u$ \textbf{ else } $u_{\textrm{max}} = u$
        \ENDFOR
        \STATE $u_\alpha = u_{\textrm{min}}$
        \STATE \textbf{if } $\aamax{k\in\K}\p{\ksdhv-\bar{K}_{k}^{\,\bullet\ceil{(B_1+1)(1-u_\alpha w_k)}}}>0$ \textbf{ then return } 1 \textbf{ else return } 0 
        \STATE \textbf{Time complexity:} $\mathcal{O}\p{\abs{\K}\p{B_1+B_2}N^2}$
    \end{algorithmic}
    \end{algorithm}

    We verify in the next proposition that performing this correction indeed ensures that our aggregated test $\dbb$ has the prescribed level $\alpha$.

    \begin{proposition}
        \label{prop:agglevel}
        For $\alpha\in(0,1)$ and a collection of kernels $\K$, the aggregated test $\dbb$ satisfies 
        $$
        \pr{\dbbv=1} \leq \alpha
        $$    
        asymptotically using a wild bootstrap (with Lipschitz continuity of $h_{p,k}$ required) and non-asymptotically using a parametric bootstrap.
    \end{proposition}

    The proof of \Cref{prop:agglevel} is presented in \Cref{proof:agglevel}.
    Details about the asymptotic result for the wild bootstrap case are reported in \Cref{sec:wilbootstrap}.
    We now provide guarantees for the power of our aggregated test \ksdagg in terms of its uniform separation rate.

    \begin{theorem}
        \label{theo:aggpower}
        Suppose the assumptions listed in \Cref{assump:aggpower} hold, and let $\psi\coloneqq p-q$.
        There exists a positive constant $C$ such that if
        $$
            \norm{\psi}_2^2 
            ~\geq~ \aamin{k\in\K} \p{ \norm{\psi-T_{\hh}\psi}_2^2
            + C \log\p{\frac{1}{\alpha w_k}}\!\frac{\sqrt{C_k}}{\beta N} }
        $$
        then the probability of type II error of $\dbb$ is controlled by $\beta$, that is,
        $
            \qr{\dbbv=0} \leq \beta
        $.
    \end{theorem}

    The proof Theorem 3.3 can be found in \Cref{proof:aggpower}, it relies on upper bounding the probability of the intersection of some events by the minimum of the probabilities of each event, and on applying Theorem 3.1 after having verified that its assumptions are satisfied for the tests with adjusted levels.
    We observe that the aggregation procedure allows to achieve the smallest uniform separation rate of the single tests $\p{\db}_{k\in\K}$ up to some logarithmic weighting term $\log(1/w_k)$.

\subsection{Bandwidth selection}
    \label{subsec:bandwidthselection}
    
    A specific application of the setting we have considered is the problem of bandwidth selection for a fixed kernel.
    Given a
    kernel $k:\R^d\times\R^d\to\R$, 
    the function
    $$
    \kl(x,y) \coloneqq k\p{\frac{x}{\lambda},\frac{y}{\lambda}}
    $$
    is also a kernel for any bandwidth $\lambda>0$.    
    A common example is the Gaussian kernel, for which we have $k(x,y)=\exp(-\norm{x-y}_2^2)$ and
    $
    \kl(x,y)=\exp\big(-{\norm{x-y}_2^2}\big/{\lambda^2}\big)
    $.
    As shown by \citet{gorham2017measuring}, a more appropriate kernel for goodness-of-fit testing using the KSD is the IMQ (inverse multiquadric) kernel, which is defined with $k(x,y)=\p{1+\norm{x-y}_2^2}^{-\beta_k}$ for a fixed parameter $\beta_k\in(0,1)$ as
    \begin{equation}
    \label{eq:imq}
    \kl(x,y) = \p{1+\frac{\norm{x-y}_2^2}{\lambda^2}}^{-\beta_k}
    \!=\ \lambda^{2\beta_k}\p{\lambda^{2}+\norm{x-y}_2^2}^{-\beta_k} 
    \!\propto\ \p{\lambda^{2}+\norm{x-y}_2^2}^{-\beta_k}
    \end{equation}
    which is the well-known form of the IMQ kernel with parameters $\lambda>0$ and $\beta_k\in(0,1)$.
    Note that it is justified to consider the kernel up to a multiplicative constant because the single and aggregated tests are invariant under this kernel transformation.

    In practice, as suggested by \citet{gretton2012kernel}, the bandwidth is often  set to a heuristic such as the median or the standard deviation of the $L^2$-distances between the samples $(X_i)_{i=1}^N$, however, these are arbitrary choices with no theoretical guarantees.
    Another common approach proposed by \citet{gretton2012optimal} for the linear-time setting, and extended to the quadratic-time setting by \citet{liu2020learning}, is to resort to data splitting in order to select a bandwidth on held-out data, by maximising for a proxy for asymptotic power (see \Cref{sec:alternativeBandwidthSelection} for details).
    Both methods were originally proposed for the two-sample problem, but extend straightforwardly to the goodness-of-fit setting.

    By considering a kernel collection $\K_\Lambda=\{k_\lambda:\lambda\in\Lambda\}$ for a collection of bandwidths $\Lambda$, we can use our aggregated test \ksdagg to test multiple bandwidths using all the data and without resorting to arbitrary heuristics. We now obtain an expression for the uniform separation rate of $\dbbl$ in terms of the bandwidths $\lambda\in\Lambda$.
    
    \begin{corollary}
        \label{cor:aggpowerbandwidth}
        Suppose the assumptions listed in \Cref{assump:aggpower} hold for $\K=\K_\Lambda=\{k_\lambda:\lambda\in\Lambda\}$, and let $\psi\coloneqq p-q$.
        There exists a positive constant $C$ such that the condition
        $$
            \norm{\psi}_2^2 
            ~\geq~ \aamin{\lambda\in\Lambda}\! \p{ \norm{\psi-T_{\hl}\psi}_2^2
            + C \log\p{\frac{1}{\alpha w_\lambda}} \frac{\sqrt{C_{\kl}}}{\beta N} }
        $$
        ensures control over the probability of type II error of the aggregated test
        $
            \qr{\dbblv=0} \leq \beta
        $.
    \end{corollary}

    \Cref{cor:aggpowerbandwidth} follows from applying \Cref{theo:aggpower} to the collection of kernels $\K_\Lambda$.
    Our results hold with great generality as we have not imposed any restrictions on $\psi\coloneqq p-q$ such as assuming it belongs to a specific regularity class.
    For this reason, the dependence on $\lambda$ in the terms $\|{\psi-T_{\hl}\psi}\|_2^2$ and $\log\!\big(1/(\alpha w_\lambda)\big) (\beta N)^{-1}\sqrt{C_{\kl}}$ is not explicit.
    In \Cref{sec:restricted_sobolev}, we characterise this dependence with regularity assumptions on $\psi\coloneqq p-q$, and derive uniform seperation rates in terms of the sample size.

\subsection{Uniform separation rates over restricted Sobolev balls}
    \label{sec:restricted_sobolev}

    In this section, we derive uniform separation rates with stronger assumptions on $p$ and $q$.
    This provides settings in which the power guaranteeing conditions of \Cref{lem:singlepower,theo:aggpower} are satisfied, and illustrates the interactions between the two terms in those conditions. 
    We make the following assumptions.
    \begin{itemize}
        \item The model density $p$ is strictly positive on its connected and compact support $S\subseteq \R^d$.
        \item The score function $\nabla \log p(x)$ is continuous and bounded on $S$.
        \item The support of the density $q$ is a connected and compact subset of $S$.
        \item The kernel used is a scaled Gaussian kernel $k_\lambda(x,y) \coloneqq \lambda^{2-d} \exp\left(-\norm{x - y}_2^2\,/\,\lambda^2\right)$.
    \end{itemize}

    In particular, any strictly positive twice-differentiable density on $\R^d$ truncated to some $d$-dimensional interval will satisfy the assumptions for the model $p$. Any density truncated to the same $d$-dimensional interval will satisfy the assumption for the density $q$.
    As an example, one can consider truncated normal densities.

    Given some smoothness parameter $s>0$, radius $R>0$ and dimension $d\in\N\setminus\{0\}$, the Sobolev ball $\Sb$ is defined as the function space
    $$
        \Sb \coloneqq\left\{ f\in L^1(\R^d)\cap L^2(\R^d) : \int_{\R^d} \norm{\xi}_2^{2s} \abs{\widehat{f}(\xi)}^2 \, \mathrm{d}\xi \leq (2\pi)^d R^2\right\},
    $$
    where $\widehat f$ denotes the Fourier transform of $f$. 
    For $s,t,d,R,L$ all strictly positive, we define the restricted Sobolev ball $\Sbt$ as containing all functions $f\in \Sb$ satisfying
    \begin{equation}
        \label{eq:sobolev_restriction}
         \int_{\norm{\xi}_2\leq t} \abs{\widehat{f}(\xi)}^2 \,\mathrm{d}\xi 
        \leq 
        \frac{1}{L}\int_{\R^d} \abs{\widehat{f}(\xi)}^2 \,\mathrm{d}\xi.
    \end{equation}
    With the Sobolev assumption $p-q\in\Sb$, the densities $p$ and $q$ can differ at any frequencies.
    The restricted Sobolev assumption $p-q\in\Sbt$ does not include the case in which the densities $p$ and $q$ differ only at low frequencies due to the additional restriction in \Cref{eq:sobolev_restriction}. 

    \begin{theorem}
        \label{theo:rate_restricted}
        Suppose the assumptions listed in \Cref{assump:rate_restricted} hold for $\K_\Lambda=\{k_\lambda:\lambda\in\Lambda\}$.
        Let $\psi\coloneqq p-q$.
        We show that there exists some $L>0$ such that if $p-q\in\Sbt$ then (i) \& (ii) hold.

        (i) Under the assumptions of \Cref{lem:singlepower}, for the KSD test with bandwidth $\lambda\coloneqq N^{-2/(4s+d)}$, the condition
        $$
            \norm{\psi}_2^2 
            ~\geq~ C N^{-4s/(4s+d)}
        $$
        for some $C>0$
        guarantees control over the probability of type II error
        $
            \qr{\dbvll=0} \leq \beta.
        $

        (ii) Under the assumptions of \Cref{theo:aggpower}, for \ksdagg  with the collection
        $$
            \Lambda \coloneqq \Big\{2^{-\ell}: \ell \in \Big\{1,\dots, \Big\lceil\frac{2}{d}\log_2\!\Big(\frac{N}{\ln(\ln(N))}\Big)\Big\rceil\Big\}\Big\}
        $$
        and weights $w_\lambda \coloneqq 6/\pi^2\ell^2$, we have 
        $
            \qr{\dbbvll=0} \leq \beta
        $
        provided that, for some $C>0$,
        $$
            \norm{\psi}_2^2 
            ~\geq~ C \p{\frac{N}{\ln\p{\ln\p{N}}}}^{-4s/(4s+d)}.
        $$
    \end{theorem}

    The proof of \Cref{theo:rate_restricted}  is presented in \Cref{proof:rate_restricted}.
    The uniform separation rate $N^{-4s/(4s+d)}$ in \Cref{theo:rate_restricted} is known to be optimal in the \emph{minimax} sense over (unrestricted) Sobolev balls $\Sb$ \citep{li2019optimality,balasubramanian2021optimality,albert2019adaptive,schrab2021mmd}.
    However, this rate is for the KSD test with bandwidth depending on the smoothness parameter $s>0$ which is unknown in practice.
    Note that even methods which split the data to perform kernel selection are not able to select this bandwidth depending on $s$ and cannot achieve the minimax rate.
    The aggregation procedure is adaptive to this unknown parameter $s$: crucially, the collection of bandwidths $\Lambda$ does not depend on $s$, and so, the aggregated \ksdagg test of \Cref{theo:rate_restricted} (ii) can be implemented in practice, unlike the single KSD test of \Cref{theo:rate_restricted} (i).
    The price to pay for this adaptivity is only an iterated logarithmic factor in the minimax rate over Sobolev balls $\Sb$.
    The uniform separation rates presented in \Cref{theo:rate_restricted} are over the restricted Sobolev balls $\Sbt$.
    Whether those rates can also be derived under less restrictive assumptions and in the more general setting of (unrestricted) Sobolev balls $\Sb$, is a challenging problem, which is left for future work.

\section{Implementation and experiments}
    \label{sec:experiments}

    We consider three different experiments based on 
    a Gamma one-dimensional distribution,
    a Gaussian-Bernoulli Restricted Boltzmann Machine, and
    a Normalizing Flow for the MNIST dataset.
    We compare our proposed aggregated test \ksdagg against three alternatives: the KSD test which uses the median bandwidth, a test which splits the data to select an `optimal' bandwidth according to a proxy for asymptotic test power, and a test which uses extra data for bandwidth selection.
    The `extra data' test is designed simply to provide a best-case scenario for the bandwidth selection procedure which maximises asymptotic test power, but it cannot be used in practice as it has access to extra data compared to the other tests.
    In order to ensure that our tests always have correct levels for all bandwidth values, dimensions and sample sizes, we use the parametric bootstrap in our experiments.

\subsection{Alternative bandwidth selection approaches}
    \label{sec:alternativeBandwidthSelection}
    \citet{gretton2012kernel} proposed to use the median heuristic as kernel bandwidth, it consists in the median of the $L^2$-distances between the samples given by
    $$
        \lambda_{\textrm{med}} \coloneqq \textrm{median}\{\norm{x_i-x_j}_2:0\leq i < j \leq N\}.
    $$
    \citet{gretton2012optimal} first proposed, for the two-sample problem using a linear-time MMD estimator, to split the data and to use half of it to select an `optimal' bandwidth which maximises a proxy for asymptotic power. 
    This procedure was extended to quadratic-time estimators and to the goodness-of-fit framework by \citet{jitkrittum2017linear}, \citet{sutherland2016generative} and \citet{liu2020learning}. These strategies rely on the asymptotic normality of the test statistic under $\HH_a$. 
    In our setting, the asymptotic power proxy to maximise is the ratio 
    \begin{equation}
        \label{eq:proxy}
        \ksdhv \ /\ \widehat\sigma_{\HH_a}
    \end{equation}
    where $\widehat\sigma_{\HH_a}^2$ is a closed-form regularised positive estimator of the asymptotic variance of $\ksdh$ under $\HH_a$ \citep[Equation 5]{liu2020learning}.
    In our experiments, we also consider a test which has access to $N$ extra samples drawn from $q$ to select an `optimal' bandwidth to run the KSD test on the original $N$ samples $\Xn$. 
    This test is interesting to compare to because it uses an `optimal' bandwidth without being detrimental to power (as it uses all $N$ samples $\Xn$ to run the test with the selected bandwidth).

\subsection{Experimental details}
\label{subsec:experimentdetails}
    Inspired from \Cref{theo:rate_restricted}, in each experiments, we use \ksdagg with different collections of bandwidths of the form
    $
    \Lambda(\ell_{-},\ell_{+}) \coloneqq \big\{2^i \lambda_{\textrm{med}} : i = \ell_{-},\dots,\ell_{+}\big\}
    $
    for the median bandwidth $\lambda_{\textrm{med}}$ and integers $\ell_{-} < \ell_{+}$ with uniform weights 
    $
    w_\lambda \coloneqq {1}/{(\ell_{+}-\ell_{-}+1)}
    $.
    For the tests which split the data, we select the bandwidth, out of the collection $\Lambda(\ell_{-},\ell_{+})$, which maximises the power proxy discussed in \Cref{sec:alternativeBandwidthSelection}.
    All our experiments are run with level $\alpha = 0.05$ using the IMQ kernel defined in \Cref{eq:imq} with parameter $\beta_k = 0.5$.
    We use a parametric bootstrap with $B_1 = B_2 = 500$ bootstrapped KSD values to compute the adjusted test thresholds, and $B_3 = 50$ steps of bisection method to estimate the correction $u_\alpha$ in \Cref{eq:ualpha}. 
    For our aggregated test, we have also designed another collection which is parameter-free (no varying parameters across experiments) and performs as well as the previous variant whose parameters are chosen appropriately for each experiments.
    The collection is obtained using the maximal inter-sample distance normalised by the dimension (see details in \Cref{sec:ksdaggstar}).
    We denote the resulting robust test by {\sc KSDAgg}$^\star$ for which we use either a wild or parametric bootstrap with $B_1=B_2=2000$ and $B_3=50$, this is the variant we recommend using in practice.
    To estimate the probability of rejecting the null hypothesis, we average the test outputs across 200 repetitions.
    All experiments have been run on an AMD Ryzen Threadripper 3960X 24 Cores 128Gb RAM CPU at 3.8GHz, the runtime is of the order of a couple of hours (significant speedup can be obtained by using parallel computing).
    We have used the \href{https://github.com/wittawatj/kernel-gof}{implementation} of \citet{jitkrittum2017linear} to sample from a Gaussian-Bernoulli Restricted Boltzmann Machine, and Phillip Lippe's \href{https://uvadlc-notebooks.readthedocs.io/en/latest/tutorial_notebooks/tutorial11/NF_image_modeling.html}{implementation} of MNIST Normalizing Flows, both under the MIT license.

    \begin{figure}[h]
    \begin{subfigure}{0.5\textwidth}
    \includegraphics[width=\linewidth]{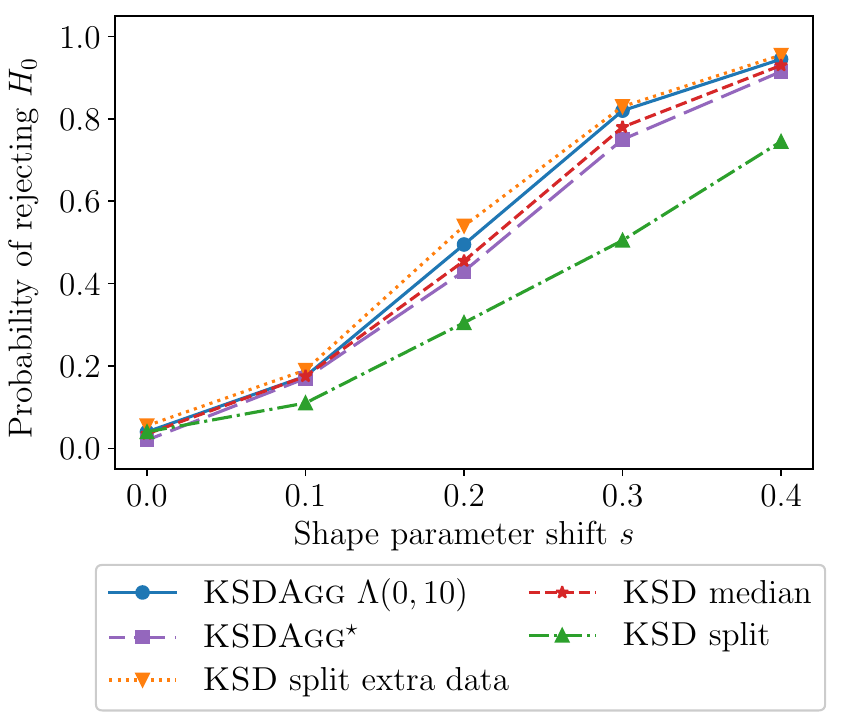}
    \caption{} 
    \label{fig:gamma}
    \end{subfigure}%
    \begin{subfigure}{0.5\textwidth}
    \includegraphics[width=\linewidth]{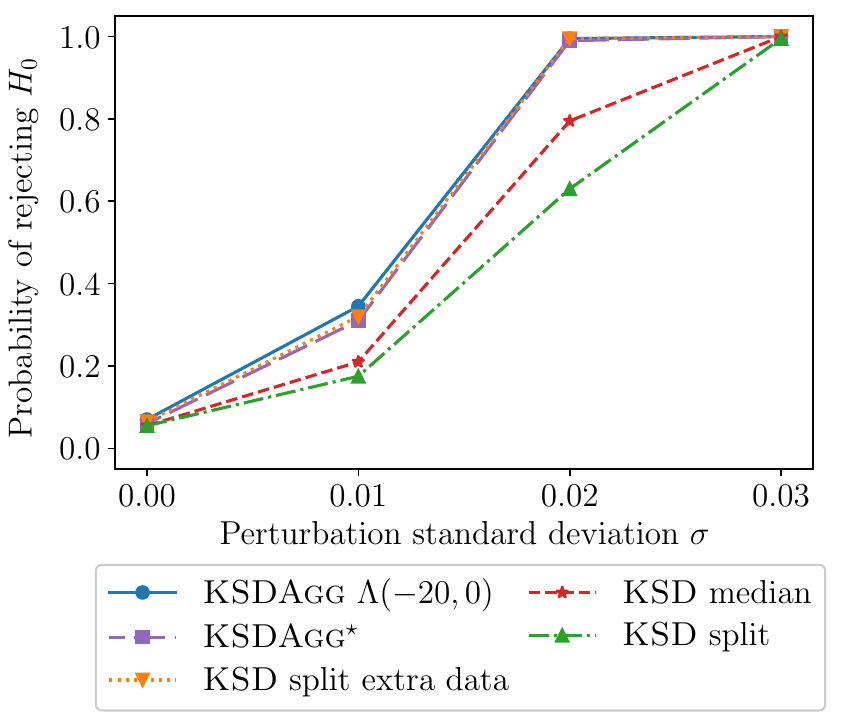}
    \caption{} 
    \label{fig:rbm}
    \end{subfigure}%
    \caption{
    (a) Gamma distribution.
    (b) Gaussian-Bernoulli Restricted Boltzmann Machine.
    }
    \end{figure}

\subsection{Gamma distribution}
\label{subsec:gamma}

    For our first experiment, we consider a one-dimensional Gamma distribution with shape parameter 5 and scale parameter 5 as the model $p$. 
    For $q$, we draw 500 samples from a Gamma distribution with the same scale parameter 5 and with a shifted shape parameter $5+s$ for $s~\!\in~\!\{0,0.1,0.2,0.3,0.4\}$. 
    We consider the collection of bandwidths $\Lambda(0,10)$.

    The results we obtained are presented in \Cref{fig:gamma}.
    We observe that all tests have the prescribed level 0.05 under the null hypothesis, which corresponds to the case $s=0$.
    As the shift parameter $s$ increases, the two densities $p$ and $q$ become more different and rejection of the null becomes an easier task, thus the test power increases.
    Our aggregated test \ksdagg achieves the same power as the KSD test with an `optimal' bandwidth selected using extra data by maximizing the proxy for asymptotic power.
    The median test obtains only slightly lower power, this closeness in power can be explained by the fact that this one-dimensional problem is a simple one.
    All four tests \ksdagg $\Lambda(0,10)$, {\sc KSDAgg}$^\star$, KSD median, and KSD split, all achieve very similar power.  
    We note that the normal splitting test has significantly lower power: this is because, even though it uses an `optimal' bandwidth, it is then run on only half the data, which results in a loss of power.

\subsection{Gaussian-Bernoulli Restricted Boltzmann Machine}
    As first considered by \citet{liu2016kernelized} for goodness-of-fit testing using the KSD, we consider a Gaussian-Bernoulli Restricted Boltzmann Machine. 
    It is a graphical model with a binary hidden variable $h\in\{-1,1\}^{d_h}$ and a continuous observable variable $x\in\R^d$. Those variables have joint density
    $$
    p(x,h) ~=~ \frac{1}{Z} \exp\p{\frac{1}{2}x^\top B h + b^\top x + c^\top h - \frac{1}{2}\norm{x}_2^2}
    $$
    where $Z$ is an unknown normalizing constant.
    By marginalising over $h$, we obtain the density $p$ of $x$
    $$
    p(x) ~=~ \sum_{h\in\{-1,1\}^{d_h}} p(x,h).
    $$
    We can sample from it using a Gibbs sampler with 2000 burn-in iterations. We use the dimensions $d=50$ and $d_h=40$ as considered by \citet{jitkrittum2017linear} and \citet{grathwohl2020learning}.
    Even though computing $p$ is intractable for large dimension $d_h$, the score function admits a convenient closed-form expression
    $$
    \nabla \log p(x) ~=~ b - x + B \frac{\exp\p{2(B^\top x + c)}-1}{\exp\p{2(B^\top x + c)}+1}.
    $$
    We draw the components of $b$ and $c$ from Gaussian standard distributions and sample Rademacher variables taking values in $\{-1,1\}$ for the elements of $B$ for the model $p$.
    We draw 1000 samples from a distribution $q$ which is constructed in a similar way as $p$ but with the difference that some Gaussian noise $\mathcal{N}(0,\sigma)$ is injected into each of the elements of $B$. 
    For the standard deviations of the perturbations, we consider $\sigma\in\{0,0.01,0.02,0.03\}$. 
    We use the collection $\Lambda(-20,0)$ (different from \Cref{subsec:gamma}), for {\sc KSDAgg}$^\star$ we use a wild bootstrap, the results are provided in \Cref{fig:rbm}.

    Again, we observe that our aggregated tests \ksdagg $\Lambda(0,20)$ and {\sc KSDAgg}$^\star$ match the power obtained by the test which uses extra data to select an `optimal' bandwidth. This means that, in this experiment, the aggregated tests obtain the same power as the `best' single test.
    The difference with the median heuristic test is significant in this experiment, and the splitting test obtains the lowest power of the four tests.
    Again, all tests have well-calibrated levels (case $\sigma=0$) and increasing the noise level $\sigma$ results in more power for all the tests.

    \begin{figure}[h]
    \begin{subfigure}{0.33\textwidth}
    \includegraphics[width=\linewidth]{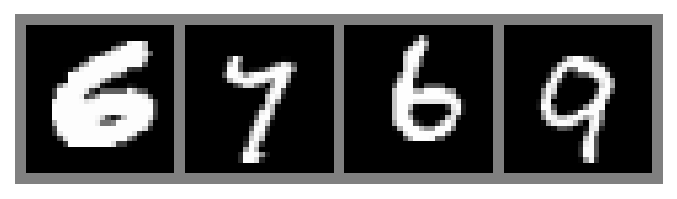}
    \caption{} 
    \end{subfigure}%
    \begin{subfigure}{0.33\textwidth}
    \includegraphics[width=\linewidth]{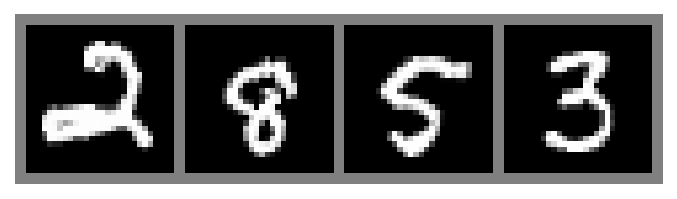}
    \caption{} 
    \end{subfigure}%
    \begin{subfigure}{0.33\textwidth}
    \includegraphics[width=\linewidth]{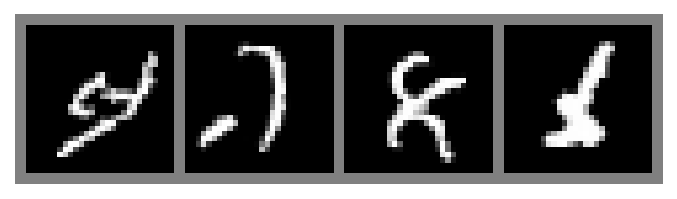}
    \caption{} 
    \end{subfigure}%
    \caption{(a) Digits from the MNIST dataset. (b, c) Digits sampled from the Normalizing Flow.}
    \label{fig:digits}
    \end{figure}

\subsection{MNIST Normalizing Flow}
\label{sec:mnistnf}
\begin{figure}[h]
    \begin{subfigure}{0.49\textwidth}
    \includegraphics[width=\linewidth]{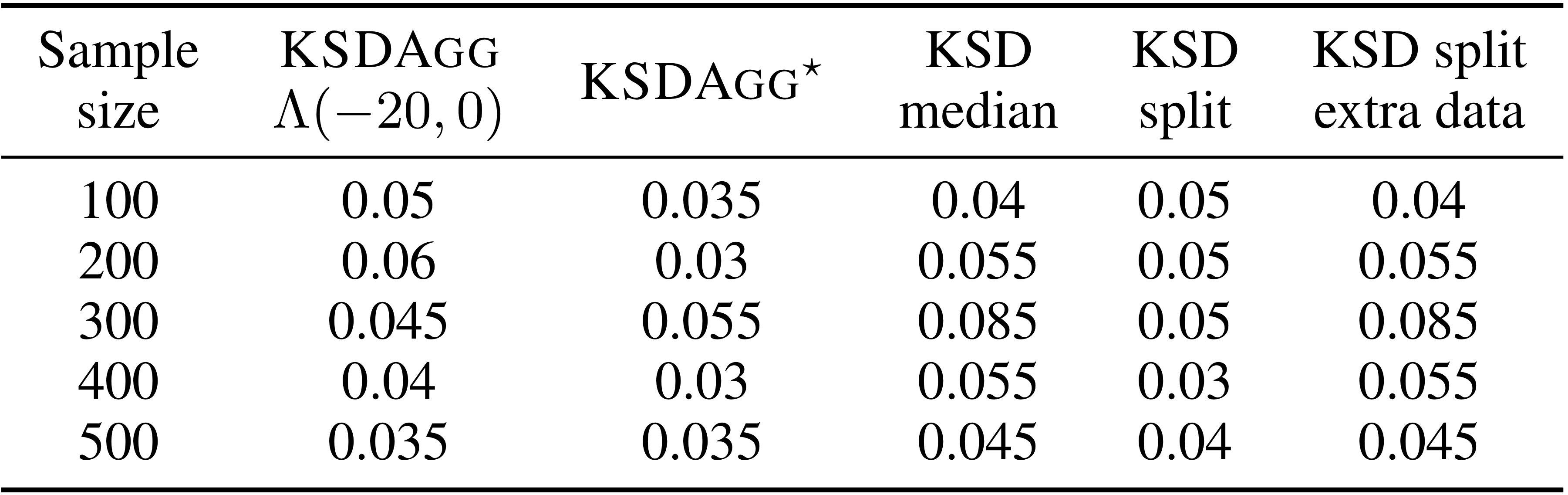}
    \caption{} 
    \label{tab:nf}
    \end{subfigure}%
    \hspace{0.02\textwidth}
    \begin{subfigure}{0.49\textwidth}
    \includegraphics[width=\linewidth]{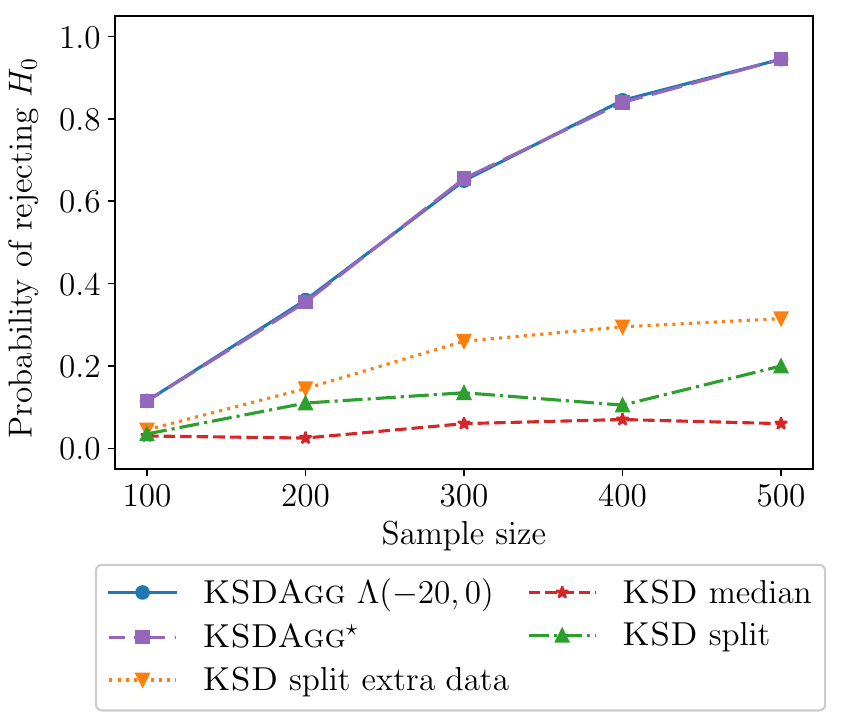}
    \caption{} 
    \label{fig:nf}
    \end{subfigure}%
    \caption{
    Normalizing Flow MNIST. (a) Level experiment. (b) Power experiment.
    }
    \end{figure}

    Finally, we consider a high-dimensional problem working with images in dimensions $28^2 = 784$. We consider a multi-scale Normalizing Flow \citep{dinh2016density,kingma2018glow} which has been trained on the MNIST dataset \citep{lecun1998gradient,lecun2010mnist}, it is a generative model which has a probability density $p$. As observed in \Cref{fig:digits}, some samples produced by the model can look exactly like MNIST digits, while others do not resemble digits. 
    This Normalizing Flow has been trained to `ideally' produce samples from the MNIST dataset. We are interested in whether or not we can detect the difference in densities.
    Given some images of digits, are we able to tell if those were sampled from the Normalizing Flow model?
    We consider the case where the samples from $q$ are drawn from the true MNIST dataset (power experiment), and the case where the images from $q$ are sampled from the Normalizing Flow model (level experiment, confirming performance for the power experiment). 
    The experiments are run with the collection of bandwidths $\Lambda(-20,0)$.
    The results are displayed in \Cref{tab:nf} and \Cref{fig:nf}.
    We use a parametric bootstrap for {\sc KSDAgg}$^\star$.

    In \Cref{tab:nf}, we observe that the four tests have correct levels (around 0.05) for the five different sample sizes considered (the small fluctuations about the designed test level can be explained by the fact that we are averaging 200 test outputs to estimate these levels).
    The well-calibrated levels obtained in \Cref{tab:nf} demonstrate the validity of the power results presented in \Cref{fig:nf}.

    As seen in \Cref{fig:nf}, only our aggregated tests \mbox{\ksdagg} and {\sc KSDAgg}$^\star$ obtain high power; they are able to detect that MNIST samples are not drawn from the Normalizing Flow. 
    The power of the other tests increases only marginally as the sample size increases. 
    The test which uses extra data to select an `optimal' bandwidth performs poorly when compared to the aggregated tests. 
    This could be explained by the fact that this test selects the bandwidth using a proxy for the asymptotic power, and that in this high-dimensional setting, the asymptotic regime is not attained with sample sizes below 500. 
    See \Cref{sec:runtimes,sec:nfdetails} for details regarding selected bandwidths and for reported runtimes.

\section{Acknowledgements}
    Antonin Schrab acknowledges support from the U.K.\ Research and Innovation (EP/S021566/1).
    Benjamin Guedj acknowledges partial support by the U.S.\ Army Research Laboratory and the U.S.\ Army Research Office, and by the U.K.\ Ministry of Defence and the U.K.\ Engineering and Physical Sciences Research Council (EP/R013616/1), and by the French National Agency for Research (ANR-18-CE40-0016-01 \& ANR-18-CE23-0015-02).
    Arthur Gretton acknowledges support from the Gatsby Charitable Foundation.
    We also thank Bharath Sriperumbudur for helpful discussions.

\clearpage

\bibliography{ksdagg_biblio}
\bibliographystyle{apalike}

\clearpage

\appendix

\section*{Supplementary material for `KSD Aggregated Goodness-of-fit Test'}

\section{Assumptions}

\subsection{Assumptions common to all theoretical results}
    \label{assump:common}
    \begin{itemize}
        \item kernel $k$ is $C_0$-universal \citep[Definition 4.1]{carmeli2010vector}
        \item $\ksdv =\E_{q,q}[\hh(X,Y)] < \infty$
        \item $C_k\geq 1$, where $C_k \coloneqq \E_{q,q}[\hh(X,Y)^2] <\infty$ as defined in \Cref{eq:ksd}
        \item $\norm{q}_\infty\leq M$ for some $M>0$
        \item $\mathbb{E}_q \Big[\norm{\nabla \p{\log \frac{p(X)}{q(X)}}}_2^2\Big]<\infty$
        \item $\alpha\in(0,e^{-1})$
        \item $\beta\in(0,1)$
    \end{itemize}

\subsection{Assumptions for \Cref{lem:singlepower}}
    \label{assump:singlepower}
    \begin{itemize}
        \item Assumptions of \Cref{assump:common}
        \item $B_1\geq 3\big(\!\log\!\big(8/\beta\big)+\alpha(1-\alpha)\big)\big/\alpha^2$
        \item assumption for parametric bootstrap: $n \big/ \sqrt{C_k} \geq \ln\p{1/\alpha}$
        \item assumption for wild bootstrap: Lipschitz continuity of $h_{p,k}$ 
        \item constant $C$ depends on $M$ and $d$
    \end{itemize}

\subsection{Assumptions for \Cref{theo:aggpower} and \Cref{cor:aggpowerbandwidth}}
    \label{assump:aggpower}
    \begin{itemize}
        \item Assumptions of \Cref{assump:common}
        \item $B_1\geq 12\Big(\underset{k\in\K}{\mathrm{max}}\, w_k^{-2}\Big)\big(\log\p{8/\beta}+\alpha(1-\alpha)\big)\big/\alpha^2$
        \item $B_2\geq 8\log\!\big(2/\beta\big)\big/\alpha^2$
        \item $B_3 \geq \log_2\!\big(4\,\aamin{k\in\K}w_k^{-1} \big/\alpha\big)$
        \item positive weights $(w_k)_{k\in\K}$ satisfying $\sum_{k\in\K} w_k \leq 1$
        \item assumption for parametric bootstrap: $n \big/ \sqrt{C_k} \geq \ln\p{1/\alpha w_k}$ for $k\in\K$
        \item assumption for wild bootstrap: Lipschitz continuity of $h_{p,k}$ 
        \item constant $C$ depends on $M$ and $d$
    \end{itemize}

\subsection{Assumptions for \Cref{theo:rate_restricted}}
    \label{assump:rate_restricted}
    \begin{itemize}
        \item Assumptions of \Cref{assump:common}
        \item $B_1\geq 12\Big(\underset{\lambda\in\Lambda}{\mathrm{max}}\, w_{\kl}^{-2}\Big)\big(\log\p{8/\beta}+\alpha(1-\alpha)\big)\big/\alpha^2$
        \item $B_2\geq 8\log\!\big(2/\beta\big)\big/\alpha^2$
        \item $B_3 \geq \log_2\!\big(4\,\aamin{\lambda\in\Lambda}w_{\kl}^{-1} \big/\alpha\big)$
        \item assumption for parametric bootstrap: $n \big/ \sqrt{C_{\kl}} \geq \ln\p{1/\alpha w_{\kl}}$ for $\lambda\in\Lambda$
        \item assumption for wild bootstrap: Lipschitz continuity of $h_{p,\kl}$ 
        \item constant $C$ is independent of the sample size $N$
	\item $\big\|T_{\hl}\psi\big\|_2 \leq \tilde C_0\big\|T_{\bkl}\psi\big\|_2$ for some $\tilde C_0>0$ with $\bkl$ as in \Cref{eq:kernelbound} 
    \end{itemize}

\section{Collection of bandwidths for {\sc KSDAgg}$^\star$}
\label{sec:ksdaggstar}

In this section, we explain how the collection for the robust test {\sc KSDAgg}$^\star$ is constructed.
First, we compute the maximal inter-sample distance
\begin{align*}
        \lambda_{\textrm{M}} &\coloneqq \max\big\{\norm{x_i-x_j}_2:0\leq i < j \leq N\big\}, \\
        \lambda_{\textrm{max}} &\coloneqq \max\big\{\lambda_{\textrm{M}}, 2\big\}.
\end{align*}
The collection of $B$ bandwidths is then defined as
    $$
        \left\{
            d^{-1}\lambda_{\textrm{max}}^{(i-1) / (B - 1)} : i = 1, \dots, B
        \right\}
    $$
where $d$ is the dimension of the samples. This collection is a discretisation of the interval $\left[d^{-1}, d^{-1}\lambda_{\textrm{max}}\right]$. 
In the limit as the number of bandwidths $B$ goes to infinity, the collection becomes the continuous interval.
In practice, we use $B=10$ bandwidths and we observed that increasing $B$ in all three experiments ($B=100, 200, \dots$) does not affect the power, that is, by using 10 bandwidths we obtain the same power as if we were to consider the aggregated test with the continuous collection $\left[d^{-1}, d^{-1}\lambda_{\textrm{max}}\right]$ of bandwidths.

\section{Background on Kernel Stein Discrepancy}
\label{sec:appendix_background}

    \textbf{Background.} Stein's methods \citep{stein1972bound} have been widely used in the machine learning and statistics communities. 
    At the heart of this field, for the Langevin Stein operator $\mathcal{A}_p$ defined as 
    $$
        (\mathcal{A}_p f)(x) \coloneqq f(x)^\top\nabla\log p(x) + \textrm{Tr}\big(\nabla f(x)\big),
    $$
    is the fact that smooth densities $p$ and $q$ are equal if and only if
    $$
        \mathbb{E}_q\big[(\mathcal{A}_p f)(x)\big] = 0
    $$
    for smooth functions $f$ vanishing at the boundaries.
    This is known as Stein's identity  \citep{stein1972bound,stein2004use}, and also holds more generally for other Stein operators.
    \citet{gorham2015measuring} utilised this identity to define Stein discrepancies as 
    $$
        \sup_{f\in\mathcal{F}}\ \mathbb{E}_q\big[f(x)^\top\nabla\log p(x) + \textrm{Tr}\big(\nabla f(x)\big)\big]
    $$
    for some space of smooth functions $\mathcal{F}$ satisfying Stein's identity.
    As proposed by \citet{oates2016control}, choosing $\mathcal{F}$ to be the unit ball in a Reproducing Kernel Hilbert Space \citep[RKHS]{azonszajn1950theory} denoted $\mathcal H$, that is $\mathcal{F} \coloneqq \{f:\Vert f \Vert_{\mathcal{H}}\leq 1\}$, we obtain the Kernel Stein Discrepancy \citep[KSD;][]{liu2016kernelized,chwialkowski2016kernel}.
    The KSD can also be expressed in terms of the Stein kernel as presented in \Cref{sec:background}.

    \textbf{Goodness-of-fit related work.}
    Stein methods have been used to construct goodness-of-fit tests in various types of data:
    directional data 
    \citep{xu2020stein}, 
    time-to-event data 
    \citep{fernandez2020kernelized}, 
    data on Riemannian manifolds 
    \citep{xu2021interpretable}, 
    conditional data 
    \citep{jitkrittum2020testing} , 
    graph data 
    \citep{weckbecker2022rkhs,xu2021stein,xu2022agrasst}, 
    functional data \citep{wynne2022spectral,wynne2022kernel},
    and generative data \citep{xu2022kernelised}. 
    We also point out the works of 
    \citet{fernandez2019maximum} 
    on an MMD-based goodness-of-fit test for censored data,
    of \citet{key2021composite} 
    on composite goodness-of-fit tests using either the MMD or KSD,
    of \citet{jitkrittum2017linear} 
    on a linear-time KSD goodness-of-fit test, 
    of \citet{gorham2017measuring} 
    on measuring sample quality with kernels and on KSD dominating weak convergence for some kernels, 
    of \citet{huggins2018random} 
    on random feature Stein discrepancies, 
    of \citet{korba2021kernel} 
    on the KSD Wasserstein gradient flow,
    of \citet{oates2019convergence} 
    on convergence rates for a class of estimators based on Stein's method,
    and of \citet{barp2019minimum} 
    on minimum Stein discrepancy estimators.
    \citet{lim2019kernel} 
    and
    \citet{kanagawa2019kernel} 
    use KSD tests for multiple model comparison and for comparing latent variable models, respectively.
    It is worth noting those score-based methods suffer from blindness to isolated components and mixing proportions \citep{wenliang2020blindness,zhang2022towards}.
    \citet{xu2021generalised,xu2022standardisation} present unified approaches for goodness-of-fit testing, and
    \citet{fernandez2022general} introduce a general framework for analysing kernel-based tests working directly with random functionals on Hilbert spaces.
    Finally, \citet{anastasiou2021stein} provide a review of recent developments based on Stein's methods.

\section{Asymptotic level for KSD test with wild bootstrap}
\label{sec:wilbootstrap}
    The proof of asymptotic level of the KSD test using a wild bootstrap was proved by \citet[Proposition 3.2]{chwialkowski2016kernel} relying on the result of \citet[Theorem 2.1]{leucht2012degenerate} and of and \citet[Lemma 5]{chwialkowski2014wild}.
    Their results for $V$-statistics can be extended to the use of $U$-statistics. 
    They proved this result for the more general non-i.i.d. case, while we need only the result for the simpler i.i.d. case.
    They proved that the difference between true quantiles and the wild bootstrap quantiles converges to zero in probability under the null hypothesis with the following dependence on $N$:
    $$
        \underset{t\in\R}{\text{sup}} \ \ \Big\vert\mathbb{P}\p{N B_N > t \mid \Xn} - \mathbb{P}\p{N K_N > t} \Big\vert
    $$
    converges to $0$ in probability, where $K_N$ is the KSD estimator of \Cref{eq:ksdh}, and $B_N$ is the wild bootstrap KSD of \Cref{eq:wildbootstrap}.
    Lipschitz continuity of the Stein kernel $h_{p,k}$ (defined in \Cref{sec:background}) is required for the result to hold.

\section{Runtimes comparison of all KSD tests considered}
\label{sec:runtimes}

    The time complexity of \ksdagg  is $\mathcal{O}\p{\abs{\K}\p{B_1+B_2}N^2}$ as provided in \Cref{alg:ksdagg}. 
    It grows linearly with the number of kernels $\abs{\K}$, quadratically with the sample size $N$, and linearly with the number of bootstrap samples $B_1+B_2$.

    \begin{table}[h]
    \caption{Runtimes (averaged over 10 runs and reported in seconds) for the Normalizing Flow MNIST experiment presented in \Cref{sec:mnistnf} using a wild bootstrap.}
    \label{tab:runtimes}
    \begin{center}
    \begin{small}
    \begin{sc}
    \begin{tabular}{ccccc}
    \toprule
    \begin{tabular}{@{}c@{}}sample \\ size\end{tabular} & \ksdagg & \begin{tabular}{@{}c@{}}KSD \\ median\end{tabular} & \begin{tabular}{@{}c@{}}KSD \\ split\end{tabular} & \begin{tabular}{@{}c@{}}KSD split \\ extra data\end{tabular} \\
    \midrule
    100 & 0.037 & 0.005 & 0.022 & 0.023 \\
    200 & 0.084 & 0.010 & 0.064 & 0.070 \\
    300 & 0.162 & 0.020 & 0.132 & 0.145 \\
    400 & 0.276 & 0.034 & 0.230 & 0.253 \\
    500 & 0.421 & 0.051 & 0.359 & 0.395 \\
    \bottomrule
    \end{tabular}
    \end{sc}
    \end{small}
    \end{center}
    \end{table}

    We report in \Cref{tab:runtimes} the runtimes of \ksdagg , KSD median, KSD split, and KSD split extra data, all using a wild bootstrap to estimate the quantiles in the MNIST Normalizing Flow experimental setting considered in \Cref{sec:mnistnf}.

    Since we consider a collection of bandwidths $\Lambda$, the time complexity of \ksdagg  is $\mathcal{O}(\abs{\Lambda} (B_1+B_2) N^2)$ and the one for median KSD is $\mathcal{O}(B_1 N^2)$ where, in the setting considered, we have $\abs{\Lambda} = 21$ and $B_1=B_2=500$. 
    While \ksdagg  takes roughly 10 times longer to run than KSD median, we could have expected a larger difference looking at the time complexities. 
    This can be explained by the fact that there are two major time-consuming steps: (i) computing the kernel matrices, and (ii) computing the wild bootstrap samples. 
    While (i) has complexity $\mathcal{O}(N^2)$ and (ii) complexity $\mathcal{O}(BN^2 + NB^2)$, the constant for step (i) is much larger than the one for step (ii) since (i) requires computing the matrix of pairwise distances while (ii) only consists in computing some matrix multiplications. 
    Note that to compute the $\abs{\Lambda}$ kernel matrices for \ksdagg , we need to compute the matrix of pairwise distances only once.

    When splitting the data, the computationally expensive step is to select the bandwidth. 
    All the $\abs{\Lambda}$ kernel matrices need to be computed, as for \ksdagg , which is the expensive step (i). 
    The KSD split test runs only slightly faster than the KSD split extra test; it runs faster than \ksdagg  but their runtimes are of the same order of magnitude.

\section{Details on MNIST Normalizing Flow experiment}
\label{sec:nfdetails}

    We provide details about the results reported in \Cref{fig:nf} of the MNIST Normalizing Flow experiment presented in \Cref{sec:mnistnf}.
    In that case, the median bandwidth $\lambda_{\textrm{med}} $ is on average $2437$. 
    The collection of bandwidths considered for \ksdagg  is $\Lambda(-20,0) \coloneqq \big\{2^i \lambda_{\textrm{med}} : i = -20,\dots,0\big\}$.
    When \ksdagg  rejects the null hypothesis, the smallest bandwidth $2^{-20} \lambda_{\textrm{med}}\approx 2^{-20}\cdot 2437 \approx 0.002$ (among others) rejects the single test with adjusted level.
    The bandwidth selected by KSD split extra data is the largest bandwidth $2^{0} \lambda_{\textrm{med}}\approx 2437$.
    The bandwidth selection is performed by maximizing the criterion in \Cref{eq:proxy}.
    \citet{sutherland2016generative} showed that this is equivalent to maximizing asymptotic power in the case of the MMD; the same result holds straightforwardly for the KSD due to similar asymptotic properties.
    The criterion only maximizes asymptotic power and has no guarantee when using limited data.
    In this high-dimensional setting ($d=784$) with sample sizes smaller than $500$, the asymptotic regime is not attained, and the criterion used for bandwidth selection does not maximize the power in this non-asymptotic setting.
    So, even though split extra has access to some extra data, it does not have an accurate criterion to select the bandwidth and ends up selecting the largest bandwidth, which is not well-adapted to this problem. 
    This explains the low power obtained by KSD split extra data.

\section{Details on our aggregation procedure}
\label{sec:aggdetails}

    \textbf{More powerful than Bonferroni correction.}
    The Bonferroni correction for multiple testing corresponds to using the adjusted level $\alpha/\abs{\K}$ for each of the $\abs{\K}$ tests.
    With the correction used for \ksdagg  with uniform weights, the adjusted level for each test is $u_\alpha/\abs{\K}$ with $u_\alpha$ defined in \Cref{eq:ualpha}.
    It can be shown that $u_\alpha\geq\alpha$ \citep[Lemma 4]{albert2019adaptive}, which means that \ksdagg  will always reject the null when the test with Bonferroni correction would reject it.
    When using uniform weights, the multiple testing correction we use is guaranteed to always lead to a test as least as powerful as the one using the Bonferroni correction.
    The proof that the Bonferroni correction guarantees control of the probability of type I error uses a loose union bound argument, the method we use aims to tighten this loose upper bound. 
    We present edge-case examples to illustrate the strengths of our multiple testing correction. 
    First, assume that the $\abs{\K}$ events are all disjoints, then the union bound is tight and both methods yield the adjusted levels $\alpha/\abs{\K}$. 
    Second, assume that all events are the same (\emph{e.g.} same kernels), then the Bonferroni correction still yields adjusted levels $\alpha/\abs{\K}$, while our multiple testing strategy can detect that there is nothing to correct for and provide `adjusted' levels $\alpha$.
    We note that, given some fixed weights and level, there does not exist a single $u_\alpha$ associated to them, it will depend on what the events are.

    \textbf{Choice of weights.}
    Without any prior knowledge (which is often the case in practice), we recommend using uniform weights since we do not expect any particular kernels/bandwidths to be better-suited than others. 
    If the user has some prior knowledge of which kernels/bandwidths would be better for the task considered, then those should be given higher weights. 
    We allow for weights whose sum is strictly smaller than 1 simply for the convenience of being able to add a new kernel/bandwidth with a new weight without changing the previous weights (as in \Cref{theo:rate_restricted} with weights $6/\pi^2\ell^2$ for $\ell\in\N\setminus\{0\}$). 
    Multiplying all the weights by a constant simply results in dividing the correction $u_\alpha$ defined in \Cref{eq:ualpha} by the same constant. 
    This means that the product $u_\alpha w_\lambda$ remains the same, and hence the definition of the aggregated test is not affected by this change. 
    For simplicity, in practice, we use weights whose sum is equal to 1.

    \textbf{Interpretability.} When \ksdagg  rejects the null hypothesis, we can check which specific kernels reject the adjusted tests: this provides the user with information about the kernels which are well-adapted to the problem considered.
    The `best' selection of kernels is naturally returned as a side-effect of the test (without requiring data splitting). 
    This contributes to the interpretability of \ksdagg , for instance in the case where different kernels prioritise different features.

    \textbf{Extension to uncountable collections of kernels.}
    Our \ksdagg  test aggregates over a finite collection of kernels.
    While even for large collections of kernels \ksdagg  retains high power (due our multiple testing correction), it would be theoretically interesting to be able to consider an uncountable/continuous collection of kernels (parametrised by the kernel bandwidth on the positive real line for example).
    Optimizing the kernel parameters continuously can be done by selecting the parameters which maximise a proxy for test power as in \Cref{eq:proxy}, but to the best of our knowledge, this has currently never been done without data splitting (which usually result in a loss of power). 
    The extension to the case of uncountable collections remains a very challenging problem, and it would be of great theoretical interest to derive such a method in the future.
    However, as shown empirically in \Cref{sec:ksdaggstar}, with a discretisation using only 10 bandwidths, our aggregated test {\sc KSDAgg}$^\star$ already obtains the same power as the test using the continuous interval (in limit of the discretisation).

    \textbf{Extension to KSD tests with adaptive features.}
    Our proposed test, \ksdagg , provides a solution to the problem of KSD adaptivity in the goodness-of-fit framework without requiring data splitting.
    A potential future direction of interest could be to tackle the adaptivity problem of the KSD-based linear-time goodness-of-fit test proposed by \citet{jitkrittum2017linear}.
    In this setting, the data is split to select feature locations (and the kernel bandwidth), the KSD test is then run using those adaptive features. 
    A challenging problem would be to obtain those adaptive features using an aggregation procedure which avoids splitting the data. 

\section{Adversarially constructed settings in which \ksdagg  might fail}
\label{sec:limitations}

    \textbf{Median bandwidth test.} In settings (possibly adversarially constructed) where the median bandwidth is the `best' bandwidth, the KSD test with median bandwidth could in theory be more competitive than \ksdagg  since by considering a large collection of bandwidths we are not only considering the `best' median bandwidth, but also `worse' bandwidths. 
    However, in practice, we cannot know in advance which bandwidth would perform well, and \ksdagg  retains power even for large collections of bandwidths (21 bandwidths considered in MNIST Normalizing Flow experiment in \Cref{sec:mnistnf}). 
    
    \textbf{Data splitting test.}
    Another setting (also possibly adversarially constructed) could be one in which the `best' bandwidth lies in between two bandwidths of our collection which are `worse' bandwidths. 
    In that case, a test which uses data splitting to select an `optimal' bandwidth would be able to select it, however, one must bear in mind the loss of power due to data splitting. 
    In such a setting, the power of \ksdagg  could then be improved by considering a finer collection of bandwidths.
    However, this issue did not arise in our experiments where we found that the aggregation procedure outperformed the competing approaches.

\section{Proofs}

\subsection{Proof of \Cref{lem:singlepower}}
\label{proof:singlepower}

    Note that 
    \begin{align*}
        \qr{\dbv~=~0}
        &~=~ \qr{\ksdhv ~\leq~ \qq}\\
        &~=~ \qr{\ksdhv - \ksdv ~\leq~ \qq - \ksdv}
    \end{align*}
    where
    $$
        \ksdv ~=~ \E_q\!\left[\ksdhv \right]
    $$
    since $\ksdh$ is an unbiased estimator.
    By Chebyshev's inequality \citep{chebyshev1899oeuvres}, we know that 
    \begin{align*}
        \beta &~\geq~ \qr{\abs{\ksdv - \ksdhv} ~\geq~ \sqrt{\frac{\mathrm{var}_q\p{\ksdhv}}{\beta}}} \\
        &~\geq~ \qr{\ksdv - \ksdhv ~\geq~ \sqrt{\frac{\mathrm{var}_q\p{\ksdhv}}{\beta}}} \\
        &~=~\qr{\ksdhv - \ksdv ~\leq~ -\sqrt{\frac{\mathrm{var}_q\p{\ksdhv}}{\beta}}}.
    \end{align*}
    We deduce that $\qr{\dbv=0}\leq \beta$ if
    \begin{align}
        \label{eq:conditionksd}
        \qq - \ksdv &~\leq~ -\sqrt{\frac{\mathrm{var}_q\p{\ksdhv}}{\beta}} \nonumber \\
        \ksdv &~\geq~ \qq + \sqrt{\frac{\mathrm{var}_q\p{\ksdhv}}{\beta}}.
    \end{align}
    The condition in \Cref{eq:conditionksd}, which controls the probability of type I error of $\db$, needs to hold with high probability since $\qq$ depends on the randomness of, either the new samples drawn from $p$ for the parametric bootstrap in \Cref{eq:parametricbootstrap}, or of the Rademacher random variables for the wild bootstrap in \Cref{eq:wildbootstrap}.
    
    We want to derive a condition in terms of $\norm{p-q}_2^2$ rather than in terms of $\ksdv$ as in \Cref{eq:conditionksd}. 
    For this, using Stein's identity, we obtain
    \begin{align*}
        \big\langle \psi, T_{\hh} \psi \big\rangle_2
        &= \int_{\R^d} \psi(y) \p{T_{\hh} \psi}\!(y) \,\dd y \\
        &= \int_{\R^d} \int_{\R^d} \hh(x,y) \psi(x) \psi(y) \,\dd x \,\dd y \\
        &= \int_{\R^d} \int_{\R^d} \hh(x,y) \p{p(x)-q(x)} \p{p(y)-q(y)} \,\dd x \,\dd y \\
        &= \int_{\R^d} \int_{\R^d} \hh(x,y) q(x) q(y) \,\dd x \,\dd y \\
        &= \E_{q,q}[\hh(X,Y)] \\
        &= \ksdv
    \end{align*}
    which gives
    $$
        \ksdv
        = \langle \psi, T_{\hh} \psi \rangle
        = \frac{1}{2}\p{\norm{\psi}_2^2+\norm{T_{\hh}\psi}_2^2 -\norm{\psi-T_{\hh}\psi}_2^2}.
    $$
    To guarantee $\qr{\dbv=0}\leq \beta$, an equivalent condition to the one presented in \Cref{eq:conditionksd} is then
    \begin{equation}
        \label{eq:conditionnorm}
        \norm{\psi}_2^2 ~\geq~ \norm{\psi-T_{\hh}\psi}_2^2 - \norm{T_{\hh}\psi}_2^2 + 2\qq + 2\sqrt{\frac{\mathrm{var}_q\p{\ksdhv}}{\beta}}
    \end{equation}
    which needs to hold with high probability over the randomness in $\qq$.
    We now upper bound the quantile and variance terms in \Cref{eq:conditionnorm} to obtain a new condition ensuring control of the type II error.

    To bound the quantile term $\qq$ using $B_1$ wild bootstrapped statistics, using the Dvoretzky--Kiefer--Wolfowitz inequality \citep{dvoretzky1956asymptotic,massart1990tight} as done by \citet[Appendix E.4]{schrab2021mmd}, it is sufficient to bound the quantile $\qqi$ using wild bootstrapped statistics without finite approximation provided that $B_1\geq \frac{3}{\alpha^2}\big(\!\log\!\big(\frac{8}{\beta}\big)+\alpha(1-\alpha)\big)$.
    Using the concentration bound for i.i.d. Rademacher chaos of \citet[Corollary 3.2.6]{victor1999decoupling} as presented by \citet[Equation 39]{kim2020minimax}, there exists some constant $C>0$ such that
    $$
        \pe{\abs{\frac{1}{N(N-1)}\sum_{1\leq i\neq j \leq N} \epsilon_i \epsilon_j \hh(X_i,X_j)}\geq t ~\Big|~ \Xn} \leq 2 \exp\p{-\frac{CtN(N-1)}{\sqrt{\displaystyle\sum_{1\leq i\neq j \leq N} \hh(X_i,X_j)^2}}}
    $$
    which gives
    $$
        \qqi ~\leq~ C \frac{\log\p{{2}/{\alpha}}}{N(N-1)} \sqrt{\displaystyle\sum_{1\leq i\neq j \leq N} \hh(X_i,X_j)^2}
    $$
    for some different constant $C>0$. 
    Using Markov's inequality, for any $\delta\in(0,1)$, with probability at least $1-\delta$ we have
    $$
        \sum_{1\leq i\neq j \leq N} \hh(X_i,X_j)^2 \leq \frac{1}{\delta} N(N-1) \E_{q,q}\!\left[\hh(X,Y)^2\right] =\frac{C_k N(N-1)}{\delta}
    $$
    which gives
    $$
        \qqi \leq C \frac{\log\p{{2}/{\alpha}}}{N(N-1)} \sqrt{\frac{C_k N(N-1)}{\delta}}
        = \frac{C}{\sqrt \delta}\frac{\log\p{{2}/{\alpha}}}{\sqrt{N(N-1)}} \sqrt{C_k}
        \leq C \log\p{\frac{1}{\alpha}} \frac{\sqrt{C_k}}{N}
    $$
    since $\alpha\in(0,e^{-1})$, where in the last inequality $C>0$ is a different constant depending on $\delta$.
    Using Dvoretzky--Kiefer--Wolfowitz inequality \citep{dvoretzky1956asymptotic} as explained above, we deduce that 
    \begin{equation}
        \label{eq:quantilebound}
        2\qq ~\leq~ 
        C \log\p{\frac{1}{\alpha}} \frac{\sqrt{C_k}}{N}
    \end{equation}
    with arbitrarily high probability, for some constant $C>0$.
    The same quantile bound also holds for the parametric bootstrap provided that $n \big/ \sqrt{C_k} \geq \ln\p{1/\alpha}$ as derived by \citet[Appendix C.4.1]{albert2019adaptive}.
    Their reasoning holds in our setting by replacing $1\big/\sqrt{\lambda_1\cdots\lambda_p\mu_1\cdots \mu_q}$ by $\sqrt{C_k}$, this is justified as for a kernel $\kl$ with bandwidths $\lambda_1,\dots,\lambda_d$ we have $\E_{q,q}[\kl(X,Y)^2] \leq C\big/\lambda_1\cdots\lambda_d$.

    We now bound the variance term in \Cref{eq:conditionksd}. Using the result of \citet[Equation 6]{albert2019adaptive}, there exists $C>0$ such that
    $$
        \textrm{var}_q\p{\ksdhv} \leq C \p{\frac{\sigma_1^2}{N} + \frac{\sigma_2^2}{N^2}}
    $$
    where
    $$
        \sigma_2^2 \coloneqq \E_{q,q}\!\left[\hh(X,Y)^2\right] = C_k
    $$
    and
    \begin{align*}
        \sigma_1^2 &\coloneqq \E_{Y\sim q}\!\left[\Big(\E_{X\sim q}\left[\hh(X,Y)\right]\Big)^2\right] \\
        &= \E_{Y\sim q}\!\left[\Big(\p{T_{\hh} \psi}\!(Y)\Big)^2\right] \\
        &= \int_{\R^d} \Big(\p{T_{\hh} \psi}\!(y)\Big)^2 q(y) \,\dd y\\
        &\leq \norm{q}_\infty \int_{\R^d} \Big(\p{T_{\hh} \psi}\!(y)\Big)^2\,\dd y \\
        &\leq M \norm{T_{\hh} \psi}_2^2
    \end{align*}
    since
    \begin{align*}
        \p{T_{\hh} \psi}\!(y) 
        &\coloneqq \int_{\R^d} \hh(x,y) \psi(x) \,\dd x \\
        &= \int_{\R^d} \hh(x,y) p(x) \,\dd x - \int_{\R^d} \hh(x,y) q(x) \,\dd x \\
        &= 0 - \int_{\R^d} \hh(x,y) q(x) \,\dd x \\
        &= - \E_{X\sim q}\left[\hh(X,y)\right]
    \end{align*}
    by Stein's identity. 
    We deduce that
    $$
        \textrm{var}_q\p{\ksdhv} \leq C \p{\frac{\norm{T_{\hh} \psi}_2^2}{N} + \frac{C_k}{N^2}}
    $$
    for some different constant $C>0$ depending on $M$ and $d$.

    Using the classical inequalities $\sqrt{x+y}\leq \sqrt x + \sqrt y$ and $2\sqrt{xy} \leq x + y$ for $x,y>0$, which are also considered in the works of \citet{fromont2013two}, \citet{albert2019adaptive} and \citet{schrab2021mmd}, we obtain
    \begin{align}
        2\sqrt{\frac{\mathrm{var}_q\p{\ksdhv}}{\beta}} 
        &\leq 2\sqrt{C \frac{\norm{T_{\hh} \psi}_2^2}{\beta N} + C\frac{C_k}{ \beta N^2}} \nonumber \\
        &\leq 2\sqrt{\norm{T_{\hh} \psi}_2^2 \frac{C}{\beta N}} + 2\sqrt{C\frac{C_k}{\beta N^2}} \nonumber\\
        &\leq \norm{T_{\hh} \psi}_2^2 + \frac{C}{\beta N} + 2\sqrt C\frac{\sqrt{C_k}}{\sqrt \beta N} \nonumber\\
        &\leq \norm{T_{\hh} \psi}_2^2 + \p{C+ 2\sqrt C} \frac{\sqrt{C_k}}{\beta N} \nonumber\\
        \label{eq:variancebound}
        &\leq \norm{T_{\hh} \psi}_2^2 + C \log\p{\frac{1}{\alpha}} \frac{\sqrt{C_k}}{\beta N}
    \end{align}
    since $\alpha\in(0,e^{-1})$, $\beta\in(0,1)$ and $C_k \geq 1$ by assumption, where the constant $C>0$ is different on the last line.
    By considering the largest of the two constants in the quantile and variance bounds of \Cref{eq:quantilebound,eq:variancebound} multiplied by two, we obtain 
    $$
        2\qq + 2\sqrt{\frac{\mathrm{var}_q\p{\ksdhv}}{\beta}} 
        ~\leq~ \norm{T_{\hh} \psi}_2^2 + C \log\p{\frac{1}{\alpha}} \frac{\sqrt{C_k}}{\beta N}.
    $$
    By combining this bound with the condition in \Cref{eq:conditionnorm}, we get that $\qr{\dbv=0}\leq \beta$ if
    \begin{align*}
        \norm{\psi}_2^2 ~&\geq~ \norm{\psi-T_{\hh}\psi}_2^2 - \norm{T_{\hh}\psi}_2^2 + \norm{T_{\hh} \psi}_2^2 + C \log\p{\frac{1}{\alpha}} \frac{\sqrt{C_k}}{\beta N}, \\
        \norm{\psi}_2^2 ~&\geq~ \norm{\psi-T_{\hh}\psi}_2^2 + C \log\p{\frac{1}{\alpha}} \frac{\sqrt{C_k}}{\beta N},
    \end{align*}
    which concludes the proof.

\subsection{Proof of \Cref{prop:agglevel}}
\label{proof:agglevel}

    Recall that the correction term in \Cref{eq:ualpha} is defined as
    \begin{equation*}
        u_\alpha
        \coloneqq \sup\bigg\{u\in\Big(0,\aamin{k\in\K}w_k^{-1}\Big): \frac{1}{B_2}\sum_{b=1}^{B_2}\one{\aamax{k\in\K}\p{\widetilde K_k^b-
            \widehat{q}_{1-u w_k}^{\,k}
        }>0} \leq \alpha\bigg\}
    \end{equation*}
    where $\widehat{q}_{1-u w_k}^{\,k} = \bar{K}_{k}^{\,\bullet\ceil{(B_1+1)(1-u w_k)}}$, as defined in \Cref{eq:quantileq}.
    Hence, we have
    \begin{equation}
        \label{eq:alphalevelsum}
        \frac{1}{B_2}\sum_{b=1}^{B_2}\one{\aamax{k\in\K}\p{\widetilde K_k^b-\widehat{q}_{1-u_\alpha w_k}^{\,k}}>0} ~\leq~ \alpha.
    \end{equation}
    Recall that our estimator is $\ksdhv$ where $\Xn \coloneqq (X_i)_{i=1}^N$ are drawn from $q$.
    Recall also from \Cref{eq:parametricbootstrap}, that for the parametric bootstrap each element $\widetilde K_k^b$ is computed by drawing new samples $(X_i')_{i=1}^{N'}$ from the model $p$ and computing $\ksdh\p{(X_i')_{i=1}^{N'}}$.
    Hence, at every sample size, under the null hypothesis $\mathcal{H}_0:p=q$ the quantities $\widetilde K_k^b$ and $\ksdhv$ are identically distributed.
    For the wild bootstrap, as defined in \Cref{eq:wildbootstrap}, \citet{chwialkowski2014wild,chwialkowski2016kernel} show that, under $\mathcal{H}_0$, $\widetilde K_k^b$ and $\ksdhv$ have the same asymptotic distribution (details are presented in \Cref{sec:wilbootstrap}).

    Therefore, by taking the expectation in \Cref{eq:alphalevelsum}, we obtain 
    \begin{align*}
        \alpha &~\geq~ \pr{\aamax{k\in\K}\p{\ksdhv-\widehat{q}_{1-u_\alpha w_k}^{\,k}}>0} \\
        &~=~ \pr{\ksdhv>\widehat{q}_{1-u_\alpha w_k}^{\,k}\ \textrm{ for some }\ k\in\K} \\
        &~=~ \pr{\Delta_{u_\alpha w_k}^{k}\!(\Xn)=1\ \textrm{ for some }\ k\in\K} \\
        &~=~ \pr{\dbbv=1}
    \end{align*}
    which holds non-asymptotically using the parametric bootstrap, and asymptotically using the wild bootstrap.
    Note that this is different from the two-sample case where using a wild bootstrap yields well-calibrated non-asymptotic levels \citep[Proposition 8]{schrab2021mmd}.
    
    The same reasoning holds by replacing $u_\alpha$ by any value $u\in(0,u_\alpha)$.
    In particular, it holds for the lower bound  of the interval obtained by performing the bisection method to approximate the supremum in the definition of $u_\alpha$.
    This lower bound is the one we use in practice, as shown in \Cref{alg:ksdagg} with the step `$u_\alpha = u_\textrm{min}$'.
    We have proved that the test with correction term $u_\alpha$ approximated with a bisection method also achieves the desired level $\alpha$.

\subsection{Proof of \Cref{theo:aggpower}}
\label{proof:aggpower}

    As explained in \Cref{proof:agglevel} and utilised in \Cref{alg:ksdagg}, we use the lower bound $\widehat u_\alpha$ of the interval obtained by performing the bisection method to approximate the supremum in the definition of $u_\alpha$ in \Cref{eq:ualpha}.
    The assumptions $B_2\geq \frac{8}{\alpha^2}\log\!\big(\frac{2}{\beta}\big)$ and $B_3 \geq \log_2\!\big(\frac{4}{\alpha}\,\aamin{k\in\K}w_k^{-1}\big)$ ensure that 
    $\widehat u_\alpha \geq \alpha / 2$ as shown by \citet[Appendix E.9]{schrab2021mmd}.

    The probability of type II error of $\dbb$ is
    \begin{align*}
        \qr{\dbbv=0}
        &~=~ \qr{\Delta_{\widehat u_\alpha w_k}^{k}\!(\Xn) = 0 \ \textrm{ for all }\ k\in\K } \\
        &~\leq~ \qr{\Delta_{\widehat u_\alpha w_{k}}^{k}\!(\Xn) = 0\ \textrm{ for some }\ k\in\K }\\
        &~\leq~ \qr{\Delta_{\alpha w_{k}\!/2}^{k}(\Xn) = 0\ \textrm{ for some }\ k\in\K }.
    \end{align*}
    To guarantee $\qr{\dbbv=0} \leq \beta$, it is then sufficient to guarantee $\qr{\Delta_{\alpha w_{k^*}\!/2}^{k^*}(\Xn) = 0} \leq \beta$ for some $k^*\in\K$ to be specified shortly in \Cref{eq:kstar}.
    By assumption, we have
    \begin{equation*}
        \label{eq:B1}
        B_1~\geq~ \p{\underset{k\in\K}{\mathrm{max}}\, w_k^{-2}} \frac{12}{\alpha^2}\p{\log\p{\frac{8}{\beta}}+\alpha(1-\alpha)}.
    \end{equation*}
    In order to apply \Cref{lem:singlepower} to $\Delta_{\alpha_k}^k$ with $\alpha_k \coloneqq \alpha w_{k}/2$ for $k\in\K$, we need to ensure that the condition on $B_1$ of \Cref{lem:singlepower} is satisfied for all $k\in\K$, that is
    $$
        B_1 ~\geq~ \frac{3}{\alpha_k^2}\p{\log\p{\frac{8}{\beta}}+\alpha_k(1-\alpha_k)}
    $$
    for all $k\in\K$.
    Since $0<\alpha_k < \alpha < e^{-1}$, we have $\alpha(1-\alpha)\geq\alpha_k(1-\alpha_k)$ for $k\in\K$.
    We also have 
    $$
        \p{\underset{k\in\K}{\mathrm{max}}\, w_k^{-2}} \frac{12}{\alpha^2} ~\geq~ 3\p{\frac{2}{\alpha w_k}}^2 = \frac{3}{\alpha_k^2}
    $$
    for all $k\in\K$.
    We deduce that
    \begin{align*}
        B_1 ~&\geq~ \p{\underset{k\in\K}{\mathrm{max}}\, w_k^{-2}} \frac{12}{\alpha^2}\p{\log\p{\frac{8}{\beta}}+\alpha(1-\alpha)} \\
        &\geq~\frac{3}{\alpha_k^2}\p{\log\p{\frac{8}{\beta}}+\alpha_k(1-\alpha_k)}
    \end{align*}
    for all $k\in\K$, and so, applying \Cref{lem:singlepower} to $\Delta_{\alpha_k}^k$ for $k\in\K$ is justified.
    We obtain that $\qr{\Delta_{\alpha w_{k}\!/2}^{k}(\Xn) = 0} \leq \beta$ if 
    $$
        \norm{\psi}_2^2 
        ~\geq~ \norm{\psi-T_{\hh}\psi}_2^2
        + C \log\p{\frac{2}{\alpha w_k}}\!\frac{\sqrt{C_k}}{\beta N},
    $$
    or, with a different constant $C>0$, if 
    $$
        \norm{\psi}_2^2 
        ~\geq~ \norm{\psi-T_{\hh}\psi}_2^2
        + C \log\p{\frac{1}{\alpha w_k}}\!\frac{\sqrt{C_k}}{\beta N}
    $$
    since 
    $\log\p{\frac{2}{\alpha w_k}} \leq \p{\log(2)+1}\log\p{\frac{1}{\alpha w_k}}$
    as $\alpha\in(0,e^{-1})$ and $w_k\in(0,1)$.
    Now, let
    \begin{equation}
        \label{eq:kstar}
        k^* \coloneqq \underset{k\in\K}{\mathrm{argmin}} \p{\norm{\psi-T_{\hh}\psi}_2^2
        + C \log\p{\frac{1}{\alpha w_k}}\!\frac{\sqrt{C_k}}{\beta N}}.
    \end{equation}
    Finally, we have $\qr{\dbbv=0} \leq \beta$ if $\qr{\Delta_{\alpha w_{k^*}\!/2}^{k^*}(\Xn) = 0} \leq \beta$, that is, if
    \begin{align*}
        \norm{\psi}_2^2 
        ~&\geq~ \norm{\psi-T_{h_{p,k^*}}\psi}_2^2
        + C \log\p{\frac{1}{\alpha w_{k^*}}}\!\frac{\sqrt{C_{k^*}}}{\beta N} \\
        &=~ \aamin{k\in\K} \p{ \norm{\psi-T_{\hh}\psi}_2^2
            + C \log\p{\frac{1}{\alpha w_k}}\!\frac{\sqrt{C_k}}{\beta N} },
    \end{align*}
    as desired.

\subsection{Proof of \Cref{theo:rate_restricted}}
\label{proof:rate_restricted}

    Recall that we suppose that the following assumptions hold.
    \begin{itemize}
        \item The model density $p$ is strictly positive on its connected and compact support $S\subseteq \R^d$.
        \item The score function $\nabla \log p(x)$ is continuous and bounded on $S$.
        \item The support of the density $q$ is a connected and compact subset of $S$.
        \item The kernel used is a scaled Gaussian kernel $k_\lambda(x,y) \coloneqq \lambda^{2-d} \exp\left(-\norm{x - y}_2^2\,/\,\lambda^2\right)$.
    \end{itemize}

    We introduce some notation. For some $c\in\R$, we write $\mathbf{c}\coloneqq(c,\dots,c)\in\R^d$. We write $(a_1,\dots, a_d) \leq (b_1,\dots,b_d)$ when $a_i\leq b_i$ for $i=1,\dots,d$. We use $C,C'$ to denote some generic constants which may change on different lines.

    Note that, by properties of compactness on $\R^d$, there exists some $a>0$ such that $S\subseteq \left[-\frac{a}{2},\frac{a}{2}\right]^d$. Since the score function $\nabla \log p(x)$ is continuous and bounded on the compact set $S$, there exists some $c_1>0$ such that 
    $$
    \abs{\left(\nabla \log p(x)\right)_i} \leq c_1 \quad\textrm{ for $i=1,\dots,d$ and for all $x\in S$}. 
    $$

    We work with a scaled Gaussian kernel with bandwidth $\lambda \leq c_2$, defined as
    $$
    k_\lambda(x,y) \coloneqq \lambda^{2-d} \exp\left(-\frac{\norm{x - y}_2^2}{\lambda^2}\right)
    $$
    which satisfies
    \begin{align*}
        \pm\nabla_x k_\lambda(x,y) &= \pm2\lambda^{-d}(y-x)\exp\left(-\frac{\norm{x - y}_2^2}{\lambda^2}\right) \leq 2\lambda^{-d} \mathbf{a}\exp\left(-\frac{\norm{x - y}_2^2}{\lambda^2}\right), \\
        \pm\nabla_y k_\lambda(x,y) &= \pm2\lambda^{-d}(x-y)\exp\left(-\frac{\norm{x - y}_2^2}{\lambda^2}\right) \leq 2\lambda^{-d} \mathbf{a}\exp\left(-\frac{\norm{x - y}_2^2}{\lambda^2}\right), \\
        \left|\sum_{i=1}^d\frac{\partial}{\partial x_i\partial y_i}\, k_\lambda(x,y)\right| &= \left|2d\lambda^{-d} - 4\lambda^{-2-d} \norm{x - y}_2^2\right|\exp\left(-\frac{\norm{x - y}_2^2}{\lambda^2}\right) \\
	    &\leq \left(2d\lambda^{-d}+4da^2\lambda^{-2-d}\right)\exp\left(-\frac{\norm{x - y}_2^2}{\lambda^2}\right).
    \end{align*}

    The Stein kernel associated to $\kl$ with $\lambda c_3\geq a$ satisfies
    \begin{align*}
        |\hl(x,y)| \leq\ 
        &\left|\p{\nabla\log p(x)^\top \nabla\log p(y)} \kl(x,y)\right|
        + \left|\nabla\log p(y)^\top \nabla_x \kl(x,y)\right| \\
        &+ \left|\nabla\log p(x)^\top \nabla_y \kl(x,y)\right| 
        + \left|\sum_{i=1}^d \frac{\partial}{\partial x_i \partial y_i}\, \kl(x,y)\right| \\
	    \leq\ &2\left(d c_1^2c_2^2 \lambda^{-d}+ 2 d c_1 a \lambda^{-d} + 3dc_3^{2}\lambda^{-d}\right)\exp\left(-\frac{\norm{x - y}_2^2}{\lambda^2}\right)\\
        \leq\ &C_0 \pi^{-d/2}\lambda^{-d}\exp\left(-\frac{\norm{x - y}_2^2}{\lambda^2}\right)
    \end{align*}
    where the constant is $C_0 \coloneqq 2 \pi^{d/2}(d c_1^2 c_2^2+ 2 d c_1 a + 3dc_3^{2})$.
    The Stein kernel can be upper bounded by a scaled Gaussian kernel on $S$.
    Writing
    $$
    \bar{k}_\lambda(x,y) \coloneqq \pi^{-d/2} \lambda^{-d} \exp\left(-\frac{\norm{x - y}_2^2}{\lambda^2}\right),
    $$
    which is of the form considered by \citet[Section 3.1]{schrab2021mmd} with equal bandwidths for each dimension, 
    we have shown that
    \begin{equation}
    \label{eq:kernelbound}
    |\hl(x,y)| \leq C_0 \bar{k}_\lambda(x,y) \qquad \text{for all $x,y\in S$.}
    \end{equation}
    Writing $\psi\coloneqq p-q$, recall from \Cref{lem:singlepower} that a sufficient condition for control of the probability of type II error by $\beta$ is 
    $$
        \norm{\psi}_2^2 
        \geq \norm{\psi-T_{\hl}\psi}_2^2
        + C \log\p{\frac{1}{\alpha}}\!\frac{\sqrt{\E_{q,q}[\hl(X,Y)^2]}}{\beta N}.
    $$
    Using the upper bound 
    $$
        \E_{q,q}[\hl(X,Y)^2]
        \leq 
        C_0^2 \E_{q,q}\left[\bar{k}_\lambda(X,Y)^2\right] 
        \leq C_0^2 \frac{M}{\lambda^d}
    $$
    where the last inequality holds as in \citet[Equation (22)]{schrab2021mmd}, we obtain the sufficient condition
    \begin{equation}
        \label{eq:conditionhalf}
        \norm{\psi}_2^2 
        \geq \norm{\psi-T_{\hl}\psi}_2^2
        + C \frac{\log\p{1/\alpha}}{\beta N\lambda^{d/2}}
    \end{equation}
    where we have absorbed the term $C_0\sqrt{M}$ in the constant $C$. 

    We would like to upper bound the term $\norm{\psi-T_{\hl}\psi}_2^2$ by $\norm{\psi-T_{\bkl}\psi}_2^2$ but this is not possible using \Cref{eq:kernelbound}. Instead, we can use the triangle inequality to get
    \begin{align*}
        \norm{\psi-T_{\hl}\psi}_2^2 
        &\leq \norm{\psi-T_{\bkl}\psi}_2^2 + \norm{T_{\bkl}\psi - T_{\hl}\psi}_2^2 \\
        &\leq \norm{\psi-T_{\bkl}\psi}_2^2 + \left(\norm{T_{\bkl}\psi}_2 + \norm{T_{\hl}\psi}_2\right)^2 \\
        &\leq \norm{\psi-T_{\bkl}\psi}_2^2 + 2\left(\norm{T_{\bkl}\psi}_2^2 + \norm{T_{\hl}\psi}_2^2\right) \\
        &\leq \norm{\psi-T_{\bkl}\psi}_2^2 + 2(\tilde C_0^2+1)\norm{T_{\bkl}\psi}_2^2
    \end{align*}
    since
    $$
    \big\|T_{\hl}\psi\big\|_2 \leq \tilde C_0\big\|T_{\bkl}\psi\big\|_2
    $$
    for some constant $\tilde C_0>0$ by assumption (see \Cref{assump:rate_restricted}).

    Recall that we assume that $\psi\in\Sbt$ for some $L$ to be determined, in particular $\psi\in\Sb$.

    For the term $\norm{\psi-T_{\bkl}\psi}_2^2$, we use the fact that $\psi\in\Sb$.
    The term $\norm{\psi-T_{\bkl}\psi}_2^2$ can then be upper bounded exactly as done by \citet[Appendix E.6]{schrab2021mmd} with the difference that we choose $\widetilde t>0$ such that $S_1<0.5$ rather than $S_1<1$ ($S_1$ is defined in \citet[Appendix E.6]{schrab2021mmd}).
    Following their reasoning, since $\psi\in\Sb$, we obtain that there exists some $S_1\in(0,0.5)$ and some constant $C>0$ (depending on $d$, $s$ and $R$) such that
    $$
    \norm{\psi-T_{\bkl}\psi}_2^2 
    \leq S_1^2 \norm{\psi}_2^2 + C \lambda^{2s}
    \leq \frac{1}{4} \norm{\psi}_2^2 + C \lambda^{2s}
    .
    $$

    To upper bound the term $\norm{T_{\bkl}\psi}_2^2$, we modify the reasoning of \citet[Appendix E.6]{schrab2021mmd} used for the term $\norm{\psi-T_{\bkl}\psi}_2^2$, and utilise the restricted Sobolev ball regularity assumption.
    First, note that since $\bkl$ is translation-invariant, $T_{\bkl}$ is a convolution as
    $$
    \big(T_{\bkl}\psi\big)(y) 
    = \int_S \bkl(x,y) \psi(x) \, \mathrm{d}x 
    = \int_S \varphi_\lambda(x-y) \psi(x) \, \mathrm{d}x 
    = \big(\psi * \varphi_\lambda\big)(y)
    $$
    where 
    $$
    \varphi_\lambda(u)\coloneqq \prod_{i=1}^d \lambda^{-1} K\left(\frac{u_i}{\lambda}\right), 
    \qquad\textrm{ and }\qquad
    K(u) \coloneqq \pi^{-1/2} \exp(-u^2).
    $$
    By properties of Fourier transforms, we have $\widehat\vl(\xi) = \prod_{i=1}^d \widehat K(\lambda \xi_i)$ for $\xi\in\R^d$.
    Note that 
    $$
    \abs{\prod_{i=1}^d \widehat K(\xi_i)}
    \leq 
    \prod_{i=1}^d \int_\R \abs{K(x)e^{-ix\xi_i}}\,\mathrm{d}x
    =
    \prod_{i=1}^d \int_\R \abs{K(x)}\,\mathrm{d}x
    = 1.
    $$
    For some $L$ to be determined, by assumption, we have $\psi\in\Sbt$, so 
    $$
    \int_{\norm{\xi}_2\leq t} \abs{\widehat{\psi}(\xi)}^2 \,\mathrm{d}\xi 
    \leq
    \frac{1}{L}
    \int_{\R^d} \abs{\widehat{\psi}(\xi)}^2 \,\mathrm{d}\xi 
    =\frac{\big \Vert \widehat\psi\big\Vert_2^2}{L}.
    $$
    For $s>0$, define
    $$
    T_s \coloneqq \sup_{\norm{\xi}_2> t} \frac{\abs{\prod_{i=1}^d \widehat{K}(\xi_i)}}{\norm{\xi}_2^s} \leq \frac{1}{t^s} < \infty.
    $$
    Then, using Plancherel's Theorem as in \citet[Appendix E.6]{schrab2021mmd}, we obtain
    \begin{align*}
        &(2\pi)^{d}\norm{T_{\bkl}\psi}_2^2 \\
        =\ &(2\pi)^{d}\norm{\psi*\vl}_2^2 \\
        =\ &\norm{\widehat\vl\widehat\psi }_2^2 \\
        =\ &\int_{\R^d} \abs{\widehat\vl(\xi)}^2\abs{\widehat\psi(\xi)}^2 \dd \xi \\
        =\ &\int_{\R^d} \left\vert\prod_{i=1}^d \widehat K(\lambda \xi_i)\right\vert^2\abs{\widehat\psi(\xi)}^2 \dd \xi \\
        =\ &\int_{\norm{\xi}_2\leq t} \left\vert\prod_{i=1}^d \widehat K(\lambda \xi_i)\right\vert^2\abs{\widehat\psi(\xi)}^2 \dd \xi + \int_{\norm{\xi}_2> t} \left\vert\prod_{i=1}^d \widehat K(\lambda \xi_i)\right\vert^2\abs{\widehat\psi(\xi)}^2 \dd \xi \\
        \leq\ &\int_{\norm{\xi}_2\leq t} \abs{\widehat\psi(\xi)}^2 \dd \xi + T_s^2\int_{\norm{\xi}_2> t} \norm{\lambda\xi}_2^{2s}\abs{\widehat\psi(\xi)}^2 \dd \xi \\
        \leq\ &\frac{1}{L}\big\Vert \widehat\psi\big\Vert_2^2 + \lambda^{2s}T_s^2\int_{\R^d} \norm{\xi}_2^{2s}\abs{\widehat\psi(\xi)}^2 \dd \xi \\
        \leq\ &\frac{(2\pi)^d}{L}\norm{\psi}_2^2 + \lambda^{2s}T_s^2 (2\pi)^d R^2.
    \end{align*}
    Combining those results, the upper bound on $\norm{\psi-T_{\hl}\psi}_2^2$ becomes
    \begin{align*}
        \norm{\psi-T_{\hl}\psi}_2^2 
        &\leq \norm{\psi-T_{\bkl}\psi}_2^2 + 2(\tilde C_0^2+1)\norm{T_{\bkl}\psi}_2^2 \\
        &\leq \p{\frac{1}{4} \norm{\psi}_2^2 + C \lambda^{2s}} + \p{\frac{2(\tilde C_0^2+1)}{L} \norm{\psi}_2^2 + C' \lambda^{2s}}\\
        &= \p{\frac{1}{4} \norm{\psi}_2^2 + C \lambda^{2s}} + \p{\frac{1}{4} \norm{\psi}_2^2 + C' \lambda^{2s}}\\
        &\leq \frac{1}{2} \norm{\psi}_2^2 + C \lambda^{2s},
    \end{align*}
    where we let $L\coloneqq 8(\tilde C_0^2+1)$.
    The condition in \Cref{eq:conditionhalf} then becomes
    $$
        \norm{\psi}_2^2 
        \geq \frac{1}{2} \norm{\psi}_2^2 + C \lambda^{2s}
        + C' \frac{\log\p{1/\alpha}}{\beta N\lambda^{d/2}}
    $$
    which gives
    \begin{equation}
    \label{eq:conditioncompact}
        \norm{\psi}_2^2 
        \geq C \p{\lambda^{2s}
        +  \frac{\log\p{1/\alpha}}{\beta N\lambda^{d/2}}}.
    \end{equation}
    Having proved the power guaranteeing condition in \Cref{eq:conditioncompact}, we can then derive rates for the KSD and \ksdagg  tests as similarly done by \citet{schrab2021mmd}.
    Set $\lambda\coloneqq N^{-2/(4s+d)}$ to get the condition
    \begin{equation*}
        \norm{\psi}_2^2 
        \geq C \p{N^{-4s/(4s+d)}
        +  N^{-1}N^{d/(4s+d)}}
        = C N^{-4s/(4s+d)}
    \end{equation*}
    which is the minimax rate over (unrestricted) Sobolev balls $\Sb$ \citep{li2019optimality,balasubramanian2021optimality,albert2019adaptive,schrab2021mmd}.

    In practice, the smoothness parameter of the Sobolev ball is not known, so we cannot set $\lambda\coloneqq N^{-2/(4s+d)}$ (\emph{i.e.} it cannot be implemented). 
    Instead, we can use our aggregated test \ksdagg .
    Adapting the proof of \Cref{theo:aggpower} with the single test power condition \Cref{eq:conditioncompact}, we obtain the aggregated test power condition
    \begin{equation}
    \label{eq:conditioncompactagg}
        \norm{\psi}_2^2 
        \geq C\, \aamin{\lambda\in\Lambda}\! \p{ \lambda^{2s}
        +  \log\p{\frac{1}{\alpha w_\lambda}}\frac{1}{\beta N\lambda^{d/2}} }.
    \end{equation}
    Similarly to \citet[Corollary 10]{schrab2021mmd}, consider
    $$
        \Lambda \coloneqq \Big\{2^{-\ell}: \ell \in \Big\{1,\dots, \Big\lceil\frac{2}{d}\log_2\!\Big(\frac{N}{\ln(\ln(N))}\Big)\Big\rceil\Big\}\Big\}, 
        \qquad
        w_\lambda \coloneqq \frac{6}{\pi^2\,\ell^2}.
    $$
    Let
    $$
    \ell_* 
    \coloneqq 
    \Big\lceil\frac{2}{4s+d}\log_2\!\Big(\frac{N}{\ln(\ln(N))}\Big)\Big\rceil
    \leq 
    \Big\lceil\frac{2}{d}\log_2\!\Big(\frac{N}{\ln(\ln(N))}\Big)\Big\rceil.
    $$
    The bandwidth $\lambda_* \coloneqq 2^{-\ell_*} \in \Lambda$ satisfies
    $$
    \ln\p{\frac{1}{w_{\lambda_*}}}
    \leq
    C \ln\p{\ell_*}
    \leq C \ln(\ln(N))
    $$
    as $w_{\lambda_*} \coloneqq 6\pi^{-2}\p{\ell_*}^{-2}$, and
    $$
    \frac{1}{2}\p{\frac{N}{\ln\p{\ln\p{N}}}}^{-2/(4s+d)}
    \leq
    \lambda_*
    \leq
    \p{\frac{N}{\ln\p{\ln\p{N}}}}^{-2/(4s+d)}.
    $$
    We get 
    $$
    \lambda_*^{2s} \leq \p{\frac{N}{\ln\p{\ln\p{N}}}}^{-4s/(4s+d)}
    $$
    and
    $$
    \log\p{\frac{1}{\alpha w_{\lambda_*}}}\frac{1}{\beta N\lambda_*^{d/2}}
    \leq C \p{\frac{N}{\ln\p{\ln\p{N}}}}^{-1} \p{\frac{N}{\ln\p{\ln\p{N}}}}^{d/(4s+d)}
    \leq C \p{\frac{N}{\ln\p{\ln\p{N}}}}^{-4s/(4s+d)}.
    $$
    The \ksdagg power condition of \Cref{eq:conditioncompactagg} then becomes 
    \begin{equation*}
        \norm{\psi}_2^2 
        \geq C \p{\frac{N}{\ln\p{\ln\p{N}}}}^{-4s/(4s+d)},
    \end{equation*}
    which concludes the proof.

\end{document}